%% file: main.tex
\documentclass[10pt,twocolumn,letterpaper]{article}

\usepackage[pagenumbers]{iccv} 

\definecolor{iccvblue}{rgb}{0.21,0.49,0.74}
\usepackage[pagebackref,breaklinks,colorlinks,allcolors=iccvblue]{hyperref}
\usepackage{caption}
\usepackage{multirow}
\usepackage{makecell}
\usepackage{bbding}
\usepackage{mdframed}
\usepackage{tabulary}
\usepackage{colortbl}
\usepackage{tikz}
\usepackage{soul}
\usepackage[hang,flushmargin]{footmisc}

\def\paperID{2972} 
\def\confName{ICCV}
\def\confYear{2025}

\definecolor{lightgray}{gray}{0.8}
\sethlcolor{lightgray}
\definecolor{color_green}{HTML}{00B050}
\definecolor{color_red}{HTML}{FF0000}
\definecolor{color_cyan}{HTML}{00B0F0}
\definecolor{color_purple}{HTML}{7030A0}

\newcommand\blfootnote[1]{%
    \begingroup 
    \renewcommand\thefootnote{}\footnote{#1}%
    \addtocounter{footnote}{-1}%
    \endgroup 
}

\title{MRGen: Segmentation Data Engine for Underrepresented MRI Modalities}

\author{Haoning Wu$^{1, 2*}$, Ziheng Zhao$^{1, 2*}$, Ya Zhang$^{1, 2}$, Yanfeng Wang$^{1, 2\dagger}$, Weidi Xie$^{1, 2\dagger}$ \\[3pt]
$^{1}$School of Artificial Intelligence, Shanghai Jiao Tong University\\[2pt]
$^{2}$Shanghai Artificial Intelligence Laboratory \\
}

\begin{document}

\twocolumn[{%
\renewcommand\twocolumn[1][]{#1}%
\maketitle
\vspace{-20pt}
\begin{center}
   \centering
   \includegraphics[width=.96\textwidth]{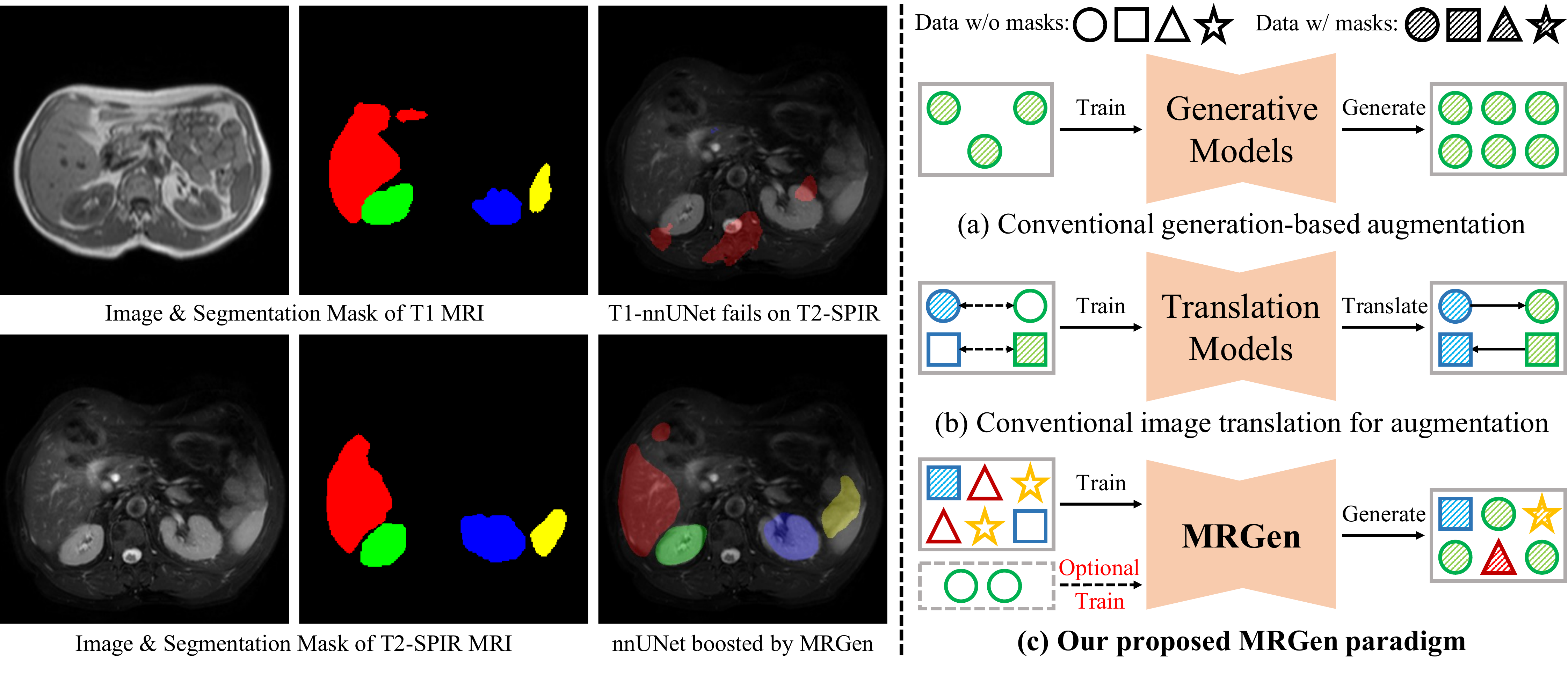}
   \vspace{-10pt}
   \captionof{figure}{
   \textbf{Motivations and Overview.}
   {\em \textbf{Left}}: 
   The heterogeneity of MRI modalities challenges the generalization of segmentation models. 
   Our proposed data engine, \textbf{MRGen}, overcomes this by controllably synthesizing training data for segmentation models.
   {\em \textbf{Right}}:
   (a) Prior generative models are restricted to data augmentation for \textbf{well-annotated modalities};
   (b) Image translation typically requires registered data pairs~(dashed lines), and is limited to specific modality conversions;
   (c) \textbf{MRGen} enables controllable generation across diverse modalities, creating data for training segmentation models towards underrepresented modalities.
   Distinct colors represent different modalities.
   }
  \label{fig:teaser}
 \end{center}
}]

\blfootnote{*: These authors contribute equally to this work. \\ 
    $\dagger$: Corresponding author.
}

\input{sec/0_abstract}    
\input{sec/1_introduction}
\input{sec/2_relatedwork}
\input{sec/3_method}
\input{sec/4_experiments}
\input{sec/5_conclusion}
\input{sec/6_acknowledgement}

{
    \small
    \bibliographystyle{ieeenat_fullname}
    \bibliography{main}
}

\input{sec/X_suppl}

\end{document}

%% file: sec/0_abstract.tex
\begin{abstract}
Training medical image segmentation models for rare yet clinically important imaging modalities is challenging due to the scarcity of annotated data, and manual mask annotations can be costly and labor-intensive to acquire. 
This paper investigates \textbf{leveraging generative models to synthesize data, for training segmentation models for underrepresented modalities}, particularly on annotation-scarce MRI. 
Concretely, our contributions are threefold:
(i) we introduce \textbf{MRGen-DB}, a large-scale radiology image-text dataset comprising extensive samples with rich metadata, including modality labels, attributes, regions, and organs information, with a subset featuring pixel-wise mask annotations;
(ii) we present \textbf{MRGen}, a diffusion-based data engine for controllable medical image synthesis, conditioned on text prompts and segmentation masks. 
MRGen can generate realistic images for diverse MRI modalities lacking mask annotations, facilitating segmentation training in low-source domains;
(iii) extensive experiments across multiple modalities demonstrate that MRGen significantly improves segmentation performance on unannotated modalities by providing high-quality synthetic data. 
We believe that our method bridges a critical gap in medical image analysis, extending segmentation capabilities to scenarios that are challenging to acquire manual annotations. 
The codes, models, and data will be publicly available at \url{https://haoningwu3639.github.io/MRGen/}.

\end{abstract}

%% file: sec/1_introduction.tex
\section{Introduction}
\label{sec:intro}
Medical image segmentation~\cite{ronneberger2015UNet, isensee2021nnUNet, MedSAM, butoi2023universeg} has shown remarkable success by training on extensive manual annotations, becoming a cornerstone of intelligent healthcare systems. 
However, developing models for underrepresented imaging modalities remains challenging, due to data privacy, modality complexity, and high cost of manual mask annotations~\cite{d2024TotalSegmentorMRI, hantze2024mrsegmentator}, especially for rare yet clinically important modalities, for example, Magnetic Resonance Imaging~(MRI). 
Despite being non-invasive and radiation-free, MRI scanning is expensive and exhibits substantial variability across modalities and scanning protocols~\cite{nananukul2024multi-source}. 
This lack of standardization and numerous hyperparameters fragment the already limited dataset, challenging the development of robust segmentation models, as illustrated in Figure~\ref{fig:teaser}~({\em left}).

In this paper, we investigate the potential of generative models, particularly diffusion models, to synthesize MRI data for training segmentation models on underrepresented modalities. While generative models offer a promising solution, they face unique challenges:
(i)~\textbf{data availability} remains a significant obstacle. 
Existing approaches have primarily focused on data augmentation for well-annotated modalities such as X-ray~\cite{bluethgen2024roentgen} and CT~\cite{guo2024maisi, hamamci2024generatect}, as depicted in Figure~\ref{fig:teaser}~({\em right}).
However, MRI data is relatively scarce, and highly diverse across modalities, making it less explored for generative modeling;
(ii)~\textbf{controllability} is critical to facilitate downstream tasks such as segmentation. 
Thus, generative models must enable controllable synthesis based on conditions such as texts, masks, or both. 
Yet, prior works~\cite{wang2024self, dorjsembe2024BrainMRISyn, meng2024MultiBrainMRI, zhan2024medm2g, konz2024anatomically} cannot simultaneously support both conditions, limiting their ability to control the generated modalities, regions, and organs effectively.

Our first contribution is to curate a high-quality, large-scale radiology image-text dataset, termed as \textbf{MRGen-DB}, which includes MRI scans across various modalities sourced from the Internet and open-source repositories. 
The dataset consists of nearly 250,000 2D slices, enriched with detailed annotations such as modality labels, attributes, region, and organ information, with a subset providing organ masks. 
This extensive collection of image-text pairs across diverse modalities forms a robust foundation for training an MRI generative model, while the availability of mask annotations facilitates more controllable and targeted synthesis.

For controllable data generation, we present \textbf{MRGen}, a diffusion-based data engine for MRI synthesis, that supports conditioning on both text prompts and segmentation masks. We employ a two-stage training strategy: 
(i) {\em text-guided pretraining} on diverse, large-scale image-text pairs, enabling the model to synthesize images across various modalities guided by templated text descriptions; 
and (ii) {\em conditioned finetuning} on mask-annotated data, facilitating controllable generation based on organ masks. 
Consequently, such a two-stage strategy allows \textbf{MRGen} to extend its controllable generation abilities towards modalities that originally do {\em not} have segmentation annotations available, thereby enabling training segmentation models for the underrepresented modalities with synthetic data.

Overall, our contributions can be summarized as follows:
(i) we explore the use of generative models for MRI synthesis across annotation-scarce modalities, facilitating training segmentation models for underrepresented modalities;
(ii) we curate \textbf{MRGen-DB}, a large-scale radiology image-text dataset, which features detailed modality labels, attributes, regions, and organs information, with a subset of organ mask annotations, providing a robust foundation for medical generative modeling;
(iii) we develop \textbf{MRGen}, a diffusion-based data engine capable of controllable generation, conditioned on templated text prompts and segmentation masks;
(iv) we conduct extensive experiments across diverse modalities, demonstrating that \textbf{MRGen} can controllably generate high-quality MR images, improving `zero-shot' segmentation performance for unannotated modalities. 
To the best of our knowledge, this work introduces the first open-source dataset curated for medical image generation, and the first generative model tailored for general annotation-scarce MRI modalities, offering a novel solution to address the scarcity of medical data and annotations.

%% file: sec/2_relatedwork.tex
\section{Related Work}

\noindent{\textbf{Generative models}} have been a research focus in computer vision for years, with GANs~\cite{goodfellow2020generative} and diffusion models~\cite{ho2020ddpm, song2020ddim} leading the advancements.
These models have found extensive applications across various tasks, including text-to-image generation~\cite{SDM, kang2023gigagan, peebles2023DiT, wu2024megafusion}, image-to-image translation~\cite{brooks2022instructpix2pix, isola2017Pix2Pix, CycleGAN2017}, artistic creation~\cite{liu2024intelligent, controlnet, ruiz2022dreambooth}, and even challenging video generation~\cite{esser2023structure, ho2022VideoDiffusion}.
Notably, CycleGAN~\cite{CycleGAN2017} employs cycle-consistency loss to facilitate image translation with unpaired data, while Stable Diffusion series~\cite{SDM, podell2023sdxl, esserSD3} efficiently produces high-resolution images in latent space, earning broad recognition.

\vspace{2pt}
\noindent{\textbf{Medical image synthesis}} aims to leverage generative models to tackle challenges such as data scarcity~\cite{liu2024radimagegan, konz2024anatomically}, biases~\cite{ktena2024generative}, and privacy concerns~\cite{koetzier2024MedSynSurvey}.
Prior works primarily focus on X-ray~\cite{bluethgen2024roentgen}, CT~\cite{guo2024maisi, hamamci2024generatect}, and brain MRI~\cite{dorjsembe2024BrainMRISyn, meng2024MultiBrainMRI, zhang2023self}, with approaches like DiffTumor~\cite{chen2024DiffTumor} and FreeTumor~\cite{wu2024freetumor} specifically targeting tumor generation to boost tumor segmentation.
While these methods have proven effectiveness in data augmentation within well-annotated training modalities and regions, they still struggle to generalize to modalities lacking manual mask annotations.
To this end, this paper investigates adopting generative models to facilitate more robust segmentation models with high-quality synthetic data of annotation-scarce modalities.

\vspace{2pt} \noindent{\textbf{Medical image segmentation}} has been a long-standing research topic, with various architectures proposed~\cite{ronneberger2015UNet, isensee2021nnUNet, hatamizadeh2021swin, zhou2023nnformer, ma2024UMamba}. 
Recently, inspired by SAM~\cite{kirillov2023sam, ravi2024sam2}, large-scale segmentation models~\cite{MedSAM,zhao2023one, du2023segvol} have been developed. However, the heterogeneity of MRI challenges the generalization of existing models, which struggle with intensity variations among diverse modalities.
Existing methods attempt to address this with image translation~\cite{sasaki2021unit, phan2023structure, kim2024adaptive} or relying on deliberately designed augmentation strategies to learn domain-invariant content~\cite{zhou2022DualNorm, hu2023devil, su2023rethinking, ouyang2022causality, xu2022adversarial, chen2023treasure}.
In this paper, we explore controllable generative models to synthesize data for segmentation training, particularly towards underrepresented modalities lacking mask annotations, thus resembling a `zero-shot' segmentation scenario.

%% file: sec/3_method.tex
\begin{figure*}[t]
  \centering
  \includegraphics[width=\textwidth]{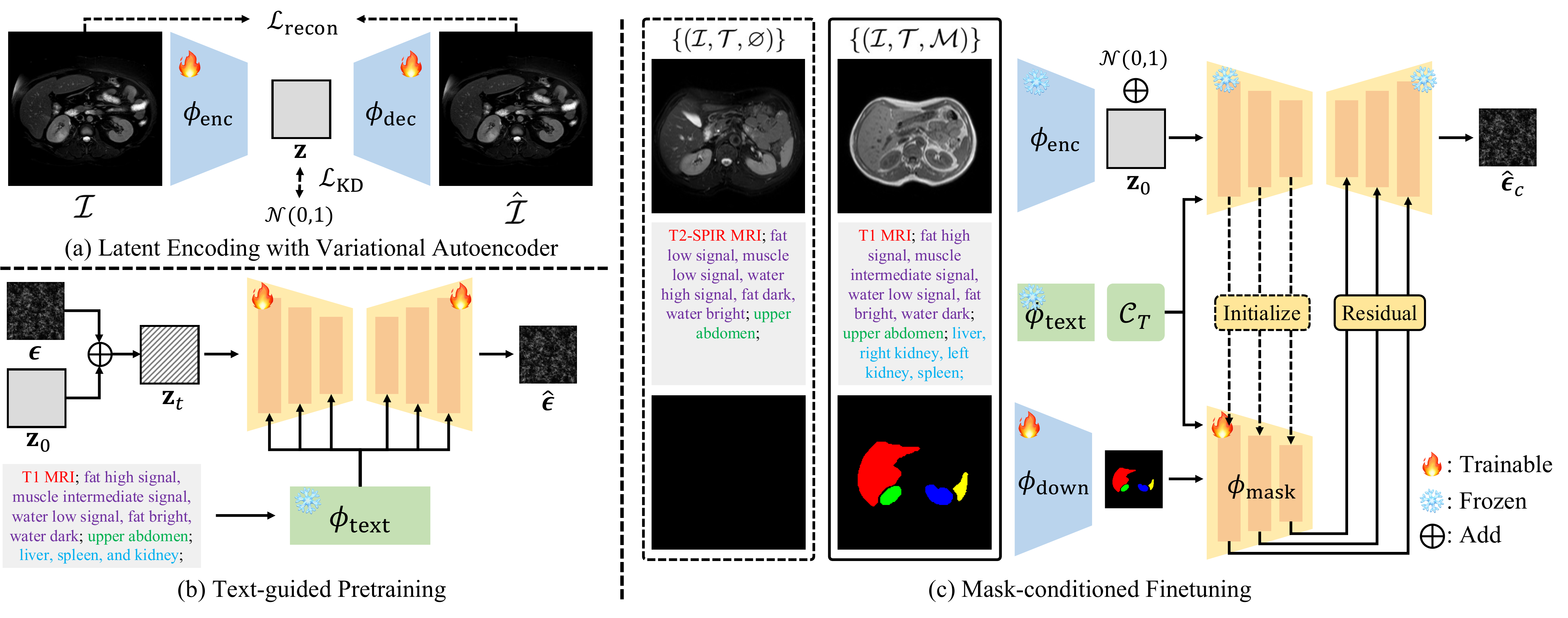} \\
  \vspace{-6pt}
  \caption{
  \textbf{Architecture Overview.} 
  Developing our MRGen involves three key steps:
  (a) Train an autoencoder on various images from MRGen-DB dataset;
  (b) Train a text-guided generative model within the latent space, using image-text pairs across diverse modalities from MRGen-DB, featuring \textcolor{color_red}{modality}, \textcolor{color_purple}{attributes}, \textcolor{color_green}{region}, and \textcolor{color_cyan}{organs} information;
  (c) Train a mask condition controller jointly on image-text pairs with and without mask annotations, enabling controllable generation based on both text prompts and masks.
  }
    \vspace{-6pt}
 \label{fig:arch}
\end{figure*}

\section{Method}
\label{sec:method}
We start to formulate the problem of interest in Sec.~\ref{sec:problem_formulation}, followed by a detailed description of the dataset curation procedure in Sec.~\ref{sec:data_curation}.
Later, we elaborate on the proposed \textbf{MRGen} architecture and the training details of our model in Sec.~\ref{sec:architecture} and Sec.~\ref{sec:model_training}, respectively. 
Lastly, we present the procedure of synthesizing and filtering samples for downstream segmentation tasks in Sec.~\ref{sec:synthetic_data_construction}.

\subsection{Problem Formulation}
\label{sec:problem_formulation}
Given a text prompt~($\mathcal{T}$) describing modality, region, and organs, along with the organs mask~($\mathcal{M}$), 
our proposed \textbf{MRGen}~($\Phi_{\mathrm{MRGen}}$) enables to generate the MR image~($\mathcal{I}$):
\begin{align*}
\mathcal{I} = \Phi_{\mathrm{MRGen}}(\mathcal{T}, \mathcal{M}; \Theta, \Theta_{c})
\end{align*}
where $\Theta$ and $\Theta_{c}$ refer to parameters of the generative model and the mask condition controller, respectively. 
Developing such a controllable data engine enables synthesizing high-quality data to train segmentation models for the challenging scenario that lacks manual annotations, for example, `zero-shot' evaluation.

\vspace{2pt} 
\noindent \textbf{Relations to existing tasks.}
Numerous studies have proven the effectiveness of generative models for data augmentation~\cite{wang2024self} on well-annotated modalities and regions, such as CT~\cite{hamamci2024generatect, guo2024maisi}, X-ray~\cite{bluethgen2024roentgen}, and brain MRI~\cite{dorjsembe2024BrainMRISyn, meng2024MultiBrainMRI}, however, this is not the focus of our work. 
Instead, we target the more challenging scenario of \textbf{synthesizing MR images for scarce and underrepresented modalities where no manual mask annotations are available}. 
While image translation methods~\cite{CycleGAN2017, kim2024unpaired, sasaki2021unit, phan2023structure} offer a potential alternative by translating richly annotated data to underrepresented modalities, these approaches often require registered data pairs for training or are limited to specific modality conversions. 
In contrast, our proposed MRGen framework offers a flexible and controllable generation pipeline, enabling the synthesis of complex abdominal MRI data across diverse modalities, even in the absence of mask annotations. 

\input{tables/dataset_statistics}

\subsection{Dataset Curation}
\label{sec:data_curation}
The scarcity of MR images with comprehensive text descriptions and mask annotations poses challenges for training generative models.
To tackle this limitation, we present \textbf{MRGen-DB}~(short for ``Database for MRI Generation''), a meticulously curated large-scale radiology dataset featuring diverse MR images enriched with modality information, detailed clinical attributes, and precise mask annotations. Below, we detail our data processing pipeline and provide comprehensive dataset statistics.

\vspace{2pt} 
\noindent \textbf{Data collection.}
The primary source of our dataset is abdominal MR images obtained from Radiopaedia\footnote{Radiopaedia.org.}, 
licensed under CC BY-NC-SA 3.0\footnote{Access to this dataset requires prior permission from Radiopaedia, after which we will provide the download link.}. 
This portion of data includes a diverse array of imaging modalities, forming extensive image-text pairs suitable for training text-guided generative models. 
Each sample consists of an MR image and its corresponding free-text modality label.

To enhance the dataset’s coverage and utility, we also augment it with abdominal MRI data from multiple open-source repositories. 
These additional data sources include modality labels and organ-specific mask annotations, forming comprehensive data triplets~($\{\mathcal{I}, \mathcal{T}, \mathcal{M}\}$). 
This augmentation enables more sophisticated controllable generation guided by both textual descriptions and anatomical masks, broadening the dataset's applicability.

\vspace{2pt} 
\noindent \textbf{Automatic annotations.}
Abdominal imaging normally exhibits significant differences across anatomical regions, such as the {\em Upper Abdominal Region} and the {\em Pelvic Region}. Relying solely on modality labels is insufficient to differentiate these distinctions.
To address this, we divide the abdomen into six anatomical regions: {\em Upper Thoracic Region}, {\em Middle Thoracic Region}, {\em Lower Thoracic Region}, {\em Upper Abdominal Region}, {\em Lower Abdominal Region}, and {\em Pelvic Region}. 
Using the pre-trained BiomedCLIP~\cite{zhang2023biomedclip}, we automatically classify all 2D slices into these regions. 
To maintain annotation quality, slices with low confidence~($<40\%$) are intentionally left unlabeled.
This process enriches the dataset with detailed region-specific information.

Distinguishing fine-grained modality differences, such as between {\em T1} and {\em T2}, presents additional challenges, even for advanced medical-specific text encoders~\cite{wu2023medklip, zhang2023knowledge}. To overcome this, we employ GPT-4~\cite{achiam2023gpt4} to map modality labels into free-text attributes that describe the signal intensities of specific tissues, including {\em fat}, {\em muscle}, and {\em water}.
For example, the {\em T1} modality can be represented as: 
{\em fat high signal, muscle intermediate signal, water low signal}. 
These detailed descriptions enable the model to understand and differentiate imaging characteristics across modalities.

To ensure the reliability of the automatic annotations, we have uniformly sampled and manually verified a data subset.
Specifically, 2\% of the region annotations and 20\% of the modality attributes have been reviewed, achieving high accuracies of 95.33\% and 91.67\%, respectively. 
This verification step ensures the high quality of the dataset and strengthens its applicability for downstream tasks.

\vspace{2pt}
\noindent \textbf{Discussion.}
After the data processing above, we assemble the \textbf{MRGen-DB} dataset, which includes about 6,000 3D volumes spanning over 100 distinct free-text MR modalities, totaling nearly 250,000 2D slices, as shown in Table~\ref{tab:dataset_statistics}. 
In the following parts, we treat each 2D slice as a data sample, which is paired with its modality label, attributes, region, and organ information, with roughly 18,000 samples featuring mask annotations. 
The scale, diversity, and fine-grained annotations of MRGen-DB offer sufficient information for training generative models tailored for MR images. 
More statistics are provided in Sec.~\ref{sec:dataset_statistics} of the \textbf{Appendix}.

\subsection{Architecture}
\label{sec:architecture}
Our research focus expects the model to leverage abundant image-text pairs and limited mask-annotated data to achieve controllable generation for underrepresented modalities. 
Specifically, our model~(\textbf{MRGen}) comprises three components: (i) latent encoding; (ii) text-guided generation; and (iii) mask-conditioned generation, as detailed below.

\vspace{2pt} 
\noindent \textbf{Latent encoding.}
To handle high-resolution medical images, we first map them into a low-dimensional latent space for efficient training.
As shown in Figure~\ref{fig:arch}~(a), we employ an autoencoder, that encodes a 2D slice~($\mathcal{I} \in \mathbb{R}^{H \times W \times 1}$) into a latent representation~($\mathbf{z} \in \mathbb{R}^{h \times w \times d}$), which can be reconstructed to image~($\hat{\mathcal{I}}$) by the decoder, expressed as: 
\begin{align*}
    \hat{\mathcal{I}} = \phi_{\mathrm{dec}}(\mathbf{z}) = \phi_{\mathrm{dec}}(\phi_{\mathrm{enc}}(\mathcal{I}))
\end{align*}

To effectively learn controllable generation based on texts and masks, the training process is carried out in two stages: 
(i) pretraining a text-guided generative model on image-text data, covering diverse modalities; 
and 
(ii) finetuning a mask condition controller jointly on data with and without mask annotations.

\vspace{2pt} 
\noindent \textbf{Text-guided generation.}
This part follows the diffusion model paradigm, comprising a forward diffusion process and a denoising process.
Concretely, the forward process progressively adds noise to the latent features~($\mathbf{z}_0$) over $T$ steps towards white Gaussian noise $\mathbf{z}_T \sim \mathcal{N} (0, 1)$.
At any intermediate timestep $t \in [1, T]$, the noisy visual features~($\mathbf{z}_t$) is expressed as:
$\mathbf{z}_t = \sqrt{\bar{\alpha}_t}\mathbf{z}_0 + \sqrt{1 - \bar{\alpha}_t} \boldsymbol{\epsilon}$, where $\boldsymbol{\epsilon} \sim \mathcal{N} (0, 1)$, and $\bar{\alpha}_t$ denotes predefined hyperparameters.

As depicted in Figure~\ref{fig:arch}~(b), the denoising process adopts a UNet~\cite{ronneberger2015UNet} and reconstructs images from noise by estimating the noise term $\hat{\boldsymbol{\epsilon}}$.
Concretely, to generate images guided by text prompts, we design a templated text prompt~($\mathcal{T}$), that consists of diverse modality labels, modality attributes, regions, and organs information, for example:
\begin{mdframed}[backgroundcolor=gray!10, linewidth=0pt]
{\em 
``T1 MRI; fat high signal, muscle intermediate signal, water low signal, fat bright, water dark; upper abdomen; liver, spleen, and kidney''.
}
\end{mdframed}
These templated prompts ensure sufficient clinical information to distinguish distinct modalities, regions, and organs.
We employ an off-the-shelf BiomedCLIP~\cite{zhang2023biomedclip} text encoder~($\phi_{\mathrm{text}}$) to encode them into embeddings, denoted as $\mathcal{C}_{T}$ = $\phi_{\mathrm{text}}(\mathcal{T})$. 
These embeddings are integrated into our model via cross-attention, serving as the key and value, with visual features~($\mathbf{z}_t$) as the query.
The output~($\mathbf{O}_{\mathrm{cross}}$) of each cross-attention layer~($\mathcal{F}_{\mathrm{cross}}$) is represented as:
\begin{align*}
    \mathbf{O}_{\mathrm{cross}} = \mathcal{F}_{\mathrm{cross}}(\mathbf{z}_t, \phi_{\mathrm{text}}(\mathcal{T}))
\end{align*}

\noindent \textbf{Mask-conditioned generation.}
We then incorporate mask conditions to enable more controllable generation. 
As presented in Figure~\ref{fig:arch}~(c), we initialize the mask encoder~($\phi_{\mathrm{mask}}$) using weights from the encoder of the diffusion UNet pre-trained in the previous stage, coupled with a learnable downsampling module~($\phi_{\mathrm{down}}$) to align dimensions. The input mask~($\mathcal{M} \in \mathbb{R}^{H\times W\times 1}$) uses distinct intensity values to represent different organs, and is integrated as a residual into the UNet decoder.
For each block~($\phi^i_{\mathrm{mask}}$) of the mask encoder, the output~($\mathbf{O}^i$) of the corresponding block~($\mathcal{F}^i$) in the diffusion UNet decoder, is formulated as:
\begin{align*}
    \mathbf{O}^i = \mathcal{F}^i(\mathbf{z}_t) + \phi_{\mathrm{mask}}^i(\mathbf{z}_t, \phi_{\mathrm{down}}(\mathcal{M}), \phi_{\mathrm{text}}(\mathcal{T}))
\end{align*}

\subsection{Model Training}
\label{sec:model_training}
Here, we present the training procedure for our proposed model, including: (i) autoencoder reconstruction, (ii) text-guided pretraining, and (iii) mask-conditioned finetuning.

\vspace{2pt} 
\noindent \textbf{Autoencoder reconstruction.}
The autoencoder for compression is trained on raw images from \textbf{MRGen-DB}, using a combination of MSE loss and KL divergence loss as follows:
$\mathcal{L}_{\mathrm{VAE}} = ||\mathcal{I} - \hat{\mathcal{I}}||_{2}^{2} + \gamma \mathcal{L}_{\mathrm{KL}}$, where $\mathcal{L}_{KL}$ imposes a KL-penalty towards a standard normal on the learned latent, similar to VAE~\cite{kingma2013VAE}, and $\gamma$ denotes a predefined weight.

\vspace{2pt}
\noindent \textbf{Text-guided pretraining.}
The diffusion-based generative model, parameterized by $\Theta$, is trained on a large number of image-text pairs, covering diverse modalities. 
The objective function is formulated as the MSE loss between the added Gaussian noise~($\boldsymbol{\epsilon}$) and the prediction~($\hat{\boldsymbol{\epsilon}}$):
\begin{align*}
    \mathcal{L} = \mathbb{E}_{t \sim [1, T], \boldsymbol{\epsilon} \sim \mathcal{N} (0, 1)} \Big[ ||\boldsymbol{\epsilon} - \hat{\boldsymbol{\epsilon}}(\mathbf{z}_t, t, \mathcal{T}) ||_{2}^{2} \Big]
\end{align*}
This pretraining phase enables \textbf{MRGen} to generate MR images across various modalities based on text prompts.

\vspace{2pt}
\noindent \textbf{Mask-conditioned finetuning.}
The mask condition controller, comprising a mask encoder~($\phi_{\mathrm{mask}}$) and a downsampling module~($\phi_{\mathrm{down}}$), is jointly trained on image-text pairs with and without mask annotations, while all other parameters~(including the autoencoder, text encoder, and diffusion UNet) remain frozen.
The training objective $\mathcal{L}_c$ is:
\begin{align*}
    \mathcal{L}_c = \mathbb{E}_{t \sim [1, T], \boldsymbol{\epsilon} \sim \mathcal{N} (0, 1)} \Big[ ||\boldsymbol{\epsilon} - \hat{\boldsymbol{\epsilon}}_{c}(\mathbf{z}_t, t, \mathcal{T}, \mathcal{M}) ||_{2}^{2} \Big]
\end{align*}
Here, incorporating data without mask annotations prevents the model from overfitting to those with masks.

\vspace{2pt} 
\noindent \textbf{Discussion.}
Such a two-stage training strategy empowers \textbf{MRGen} to achieve controllable generation across diverse modalities, 
even for those lacking mask annotations, driven by two key factors:
(i) text-guided pretraining on large-scale image-text data of various modalities equips \textbf{MRGen} with the foundational knowledge to synthesize diverse MR images based on text prompts;
(ii) mask-conditioned finetuning on partial annotated data instructs \textbf{MRGen} to integrate text and mask conditions, enabling controllability that generalizes to modalities included during pretraining.

\begin{figure}[t]
  \centering
  \includegraphics[width=0.46\textwidth]{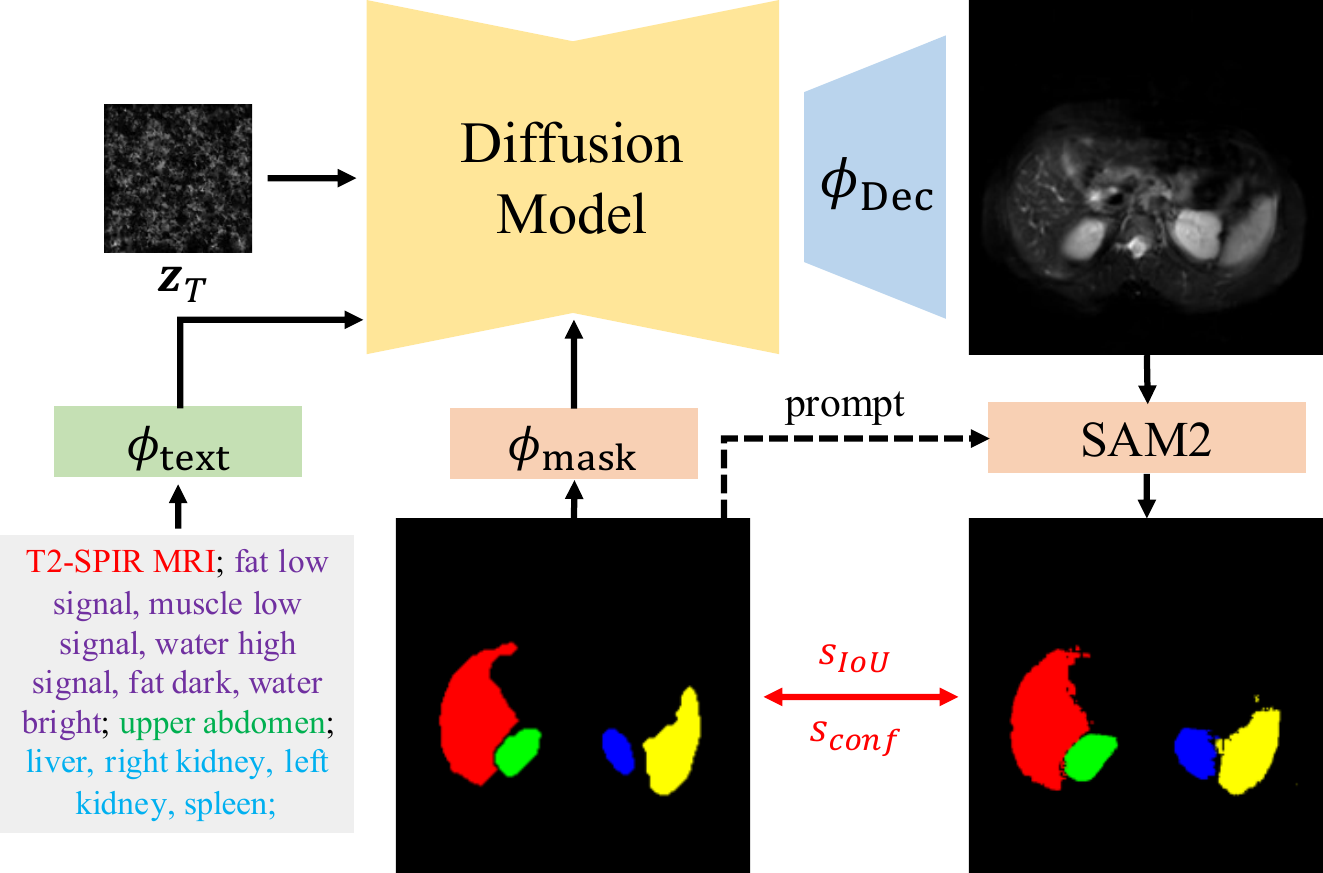} \\
  \caption{
  \textbf{Synthetic Data Construction Pipeline.}
    MRGen takes a text prompt and a mask as conditions for controllably generating MR images and employs a pretrained SAM2 model for automatic filtering to guarantee the quality of generated samples.
   }
  \vspace{-9pt}
  \label{fig:synthesize}
\end{figure}

\subsection{Synthetic Data for Segmentation Training}
\label{sec:synthetic_data_construction}

With our data engine, we can then produce MR samples for training downstream segmentation models.
At inference time, the text prompt~($\mathcal{T'}$) and controlling
organ mask~($\mathcal{M'}$) are fed into our generative model~($\Phi_{\mathrm{MRGen}}$) as conditions to generate the corresponding MR sample~($\mathcal{I'}$). 
To ensure the fidelity of the generated images on mask conditions, we design an automatic filtering pipeline using the off-the-shelf SAM2-Large~\cite{ravi2024sam2} model, as depicted in Figure~\ref{fig:synthesize}.
Specifically, we feed the conditional mask~($\mathcal{M'}$) and the generated image~($\mathcal{I'}$) into SAM2 to predict a segmentation map with a confidence score~($s_{\mathrm{conf}}$), which is used to calculate the IoU score~($s_{\mathrm{IoU}}$) against $\mathcal{M'}$.
A sample is considered to be high-quality and aligned with the mask condition if both its IoU score~($s_{\mathrm{IoU}}$) and confidence score~($s_{\mathrm{conf}}$) exceed the predefined thresholds; otherwise, it is discarded.

%% file: tables/dataset_statistics.tex
\begin{table}[t]
    \centering
    \small
    \setlength{\tabcolsep}{0.25cm} 
    \renewcommand{\arraystretch}{1.1} 
    \begin{tabular}{lcccc}
    \toprule
    Dataset & \#~Volumes & \#~Slices & \#~Masks \\
    \midrule
    Radiopaedia-MRI & 5,414 & 205,039 & ---  \\
    \midrule
    PanSeg~\cite{zhang2024PanSeg} & 766 & 33,360 & 13,779  \\
    MSD-Prostate~\cite{antonelli2022MSD} & 64 & 1,204 & 366    \\
    CHAOS-MRI~\cite{kavur2021chaos} & 60 & 1,917 & 1,492  \\
    PROMISE12~\cite{litjens2014PROMISE12} & 50 & 1,377 & 778  \\
    LiQA~\cite{liu2024LiQA} & 30 & 2,185 & 1,446  \\
    \midrule
    \textbf{Total} & \textbf{6,384} & \textbf{245,082} & \textbf{17,861} \\
    \bottomrule
    \end{tabular}
    \vspace{-3pt}
    \caption{
    \textbf{Dataset Statistics of MRGen-DB.}
    }
    \label{tab:dataset_statistics}
    \vspace{-12pt}
\end{table}



%% file: sec/4_experiments.tex
\section{Experiments}
Here, we first outline the experimental settings in Sec.~\ref{sec:experimental_settings}, followed by a comprehensive evaluation from both quantitative and qualitative perspectives in Sec.~\ref{sec:quantitative_results} and Sec.~\ref{sec:qualitative_results}. 
Lastly, we present ablation study results in Sec.~\ref{sec:ablation_studies} to prove the effectiveness of our proposed method.

\subsection{Experimental Settings}
\label{sec:experimental_settings}
Unlike existing work focusing on data augmentation for well-annotated modalities, we explore the \textbf{more challenging} scenario where certain modalities lack annotations entirely, and aim to adopt generative models to synthesize data for training segmentation models for such underrepresented modalities. 
Concretely, to simulate such a clinical scenario, we construct 5 cross-modality dataset pairs within MRGen-DB, each pair comprises a \textbf{mask-annotated source-domain} dataset~($\mathcal{D}_s$) and an \textbf{unannotated target-domain} dataset~($\mathcal{D}_t$). 
We leverage models trained on each dataset pair to generate target-domain samples for training segmentation models and assess our data engine from two aspects: 
(i) image generation quality and (ii) segmentation performance on the target-domain test set. 
Here, the test data for segmentation is strictly not used for training our generative data engines to prevent information leakage.

\input{tables/quantitative_results_generation}

\vspace{2pt} 
\noindent{\textbf{Baselines.}} 
For \textbf{generation}, we compare generated images from MRGen against three representative methods:
CycleGAN~\cite{CycleGAN2017} and UNSB~\cite{kim2024unpaired} for translating source-domain images to the target domain; 
and 
DualNorm~\cite{zhou2022DualNorm} for exhaustive augmentation of source-domain images.
For \textbf{segmentation}, we assess models trained on five data sources:
(i) source-domain data, as a baseline;
(ii) DualNorm augmented data;
(iii) CycleGAN translated data;
(iv) UNSB translated data;
and 
(v) MRGen generated data.
We adopt nnUNet~\cite{isensee2021nnUNet} and UMamba~\cite{ma2024UMamba} as segmentation frameworks for all settings, except for DualNorm, which employs a customized UNet following their official codes.
More baseline results are provided in Sec.~\ref{sec:more_quantitative_results} of the \textbf{Appendix}.

\vspace{2pt}
\noindent{\textbf{Evaluation metrics.}}
For image generation, we leverage Fréchet Inception Distance~(FID)~\cite{heusel2017FID} score to assess the diversity and quality of generated images.
For segmentation models, we employ the commonly used Dice Similarity Coefficient (DSC)~\cite{maier2018Dice} score to compare predicted masks with ground truth.
Considering segmentation consistency, we stack slices into 3D volumes, calculate the DSC for each organ individually, and average them as the final result.

\vspace{2pt}
\noindent{\textbf{Implementation details.}}
All images are resized to 512 $\times$ 512, and training is conducted on $8\times$ Nvidia A100 GPUs using the AdamW~\cite{loshchilov2018AdamW} optimizer.
We start by training the autoencoder with a learning rate of $5\times 10^{-5}$ and a batch size of $256$ for 50K iterations.
Next, the text-guided generative model and mask condition controller are trained with a learning rate of $1 \times 10^{-5}$, using batch sizes of 256 and 128 for 200K and 40K iterations, respectively.
Moreover, we randomly drop text prompts with a 10\% probability to enable classifier-free guidance~\cite{ho2022classifier}.
The compression ratio, latent dimension~$d$, KL loss weight~$\gamma$, and diffusion timesteps $T$ are set to $8$, $16$, $1\times 10^{-4}$, and $1000$, respectively.
During inference, we perform 50-step sampling using DDIM~\cite{song2020ddim}, with classifier-free guidance weight $w = 7.0$.
For each mask, we generate 20 image candidates and select the best two satisfying the predefined thresholds, which are set to 0.80 and 0.90 for IoU and confidence scores, respectively.
Conditions for target-domain data synthesis are directly derived from the region, organs information, and segmentation masks of the source-domain data.

\input{tables/quantitative_results_segmentation}

\begin{figure}[t]
  \centering
  \includegraphics[width=.47\textwidth]{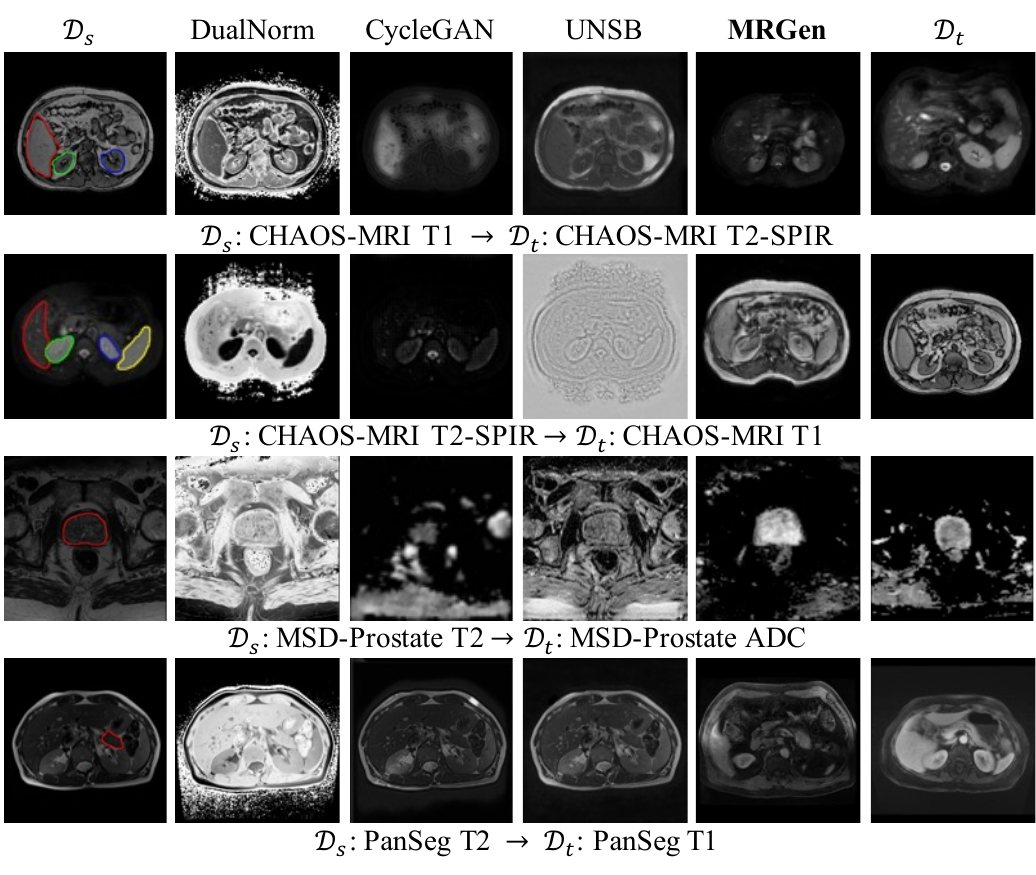} \\
  \vspace{-6pt}
  \caption{
  \textbf{Qualitative Results of Controllable Generation.} 
  We present images from source domains~($\mathcal{D}_s$) and target domains~($\mathcal{D}_t$) for reference.
  Here, \textcolor{red}{liver}, \textcolor{green}{right kidney}, \textcolor{blue}{left kidney}, \textcolor{yellow}{spleen}, \textcolor{red}{prostate},  and \textcolor{red}{pancreas} are contoured with different colors.
  }
    \vspace{-9pt}
 \label{fig:qualitative_results_generation}
\end{figure}

\subsection{Quantitative Results}
\label{sec:quantitative_results}

\noindent{\textbf{Generation.}}
As shown in Table~\ref{tab:quantitative_results_generation}, source-domain images exhibit high FID values compared to the target domain, indicating substantial discrepancies across distinct modalities.
Similarly, images augmented by DualNorm and translated by CycleGAN and UNSB also show high FID values, confirming their limited ability to emulate target-domain images. 
In contrast, our MRGen presents a significantly lower FID, demonstrating its ability to accurately generate images of target modalities, providing a foundation for training segmentation models with high-quality synthetic data.

\vspace{2pt} 
\noindent{\textbf{Segmentation.}}
As depicted in Table~\ref{tab:quantitative_results_segmentation}, we draw the following observations:
(i) notable discrepancies among different modalities challenge the generalization ability of nnUNet and UMamba trained solely on source-domain data, leading to lower average DSC scores;
(ii) DualNorm and segmentation models trained with data translated by CycleGAN and UNSB, achieve slight or moderate improvements, but consistently underperform across most scenarios;
(iii) conversely, our MRGen produces high-quality target-domain samples for training segmentation models, thus achieving the best DSC score in 8 out of 10 experiments, significantly outperforming others.
Notably, MRGen consistently improves the performance of nnUNet and UMamba, exhibiting the robustness and adaptability of its synthetic data across various segmentation architectures.
More comparisons are included in Sec.~\ref{sec:more_quantitative_results} of the \textbf{Appendix}.

\subsection{Qualitative Results}
\label{sec:qualitative_results}
\noindent \textbf{Generation.}
As presented in Figure~\ref{fig:qualitative_results_generation}, images of distinct modalities exhibit substantial visual discrepancies, making it challenging for DualNorm to simulate via exhaustive augmentation. While CycleGAN and UNSB preserve contours well, they suffer from unstable training and model collapse when learning complex transformations, thus failing to synthesize target-domain images accurately. In contrast, MRGen effectively generates images that closely resemble target domains and align with given organ masks, providing compelling evidence for controllable MRI data synthesis.

\begin{figure*}[t]
  \centering
  \includegraphics[width=.98\textwidth]{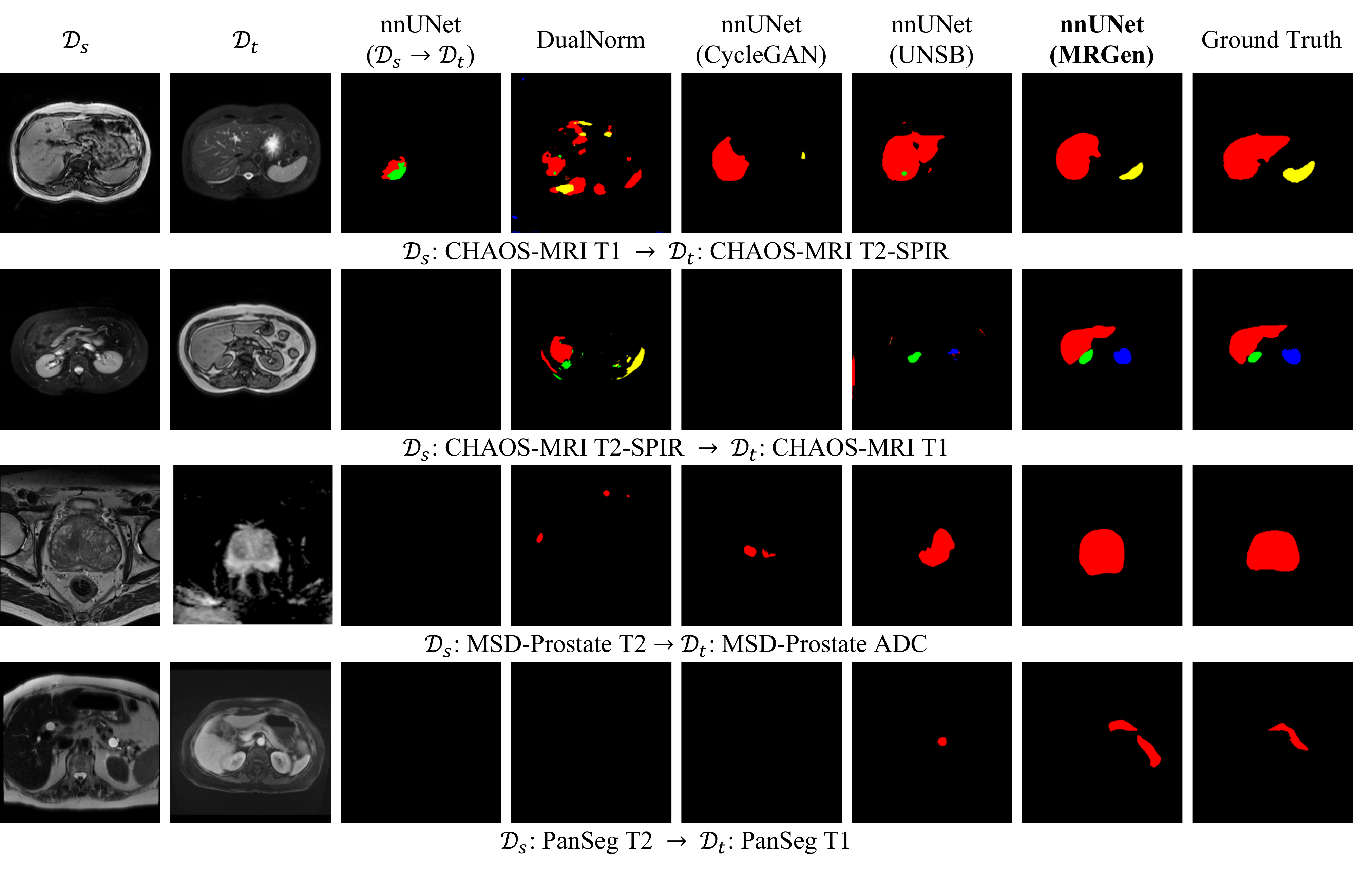} \\
  \vspace{-10pt}
  \caption{
  \textbf{Qualitative Results of Segmentation towards Unannotated Modalities.} 
  Significant imaging differences between source-domain~($\mathcal{D}_s$) and target-domain~($\mathcal{D}_t$) make segmentation on target domains~($\mathcal{D}_t$) extremely challenging.
  We visualize \textcolor{red}{liver}, \textcolor{green}{right kidney}, \textcolor{blue}{left kidney}, and \textcolor{yellow}{spleen} in the first two rows, \textcolor{red}{prostate} in the third row, and \textcolor{red}{pancreas} in the fourth row with distinct colors.
  }
    \vspace{-9pt}
 \label{fig:qualitative_results_segmentation}
\end{figure*}

\vspace{2pt}
\noindent \textbf{Segmentation.}
As illustrated in Figure~\ref{fig:qualitative_results_segmentation}, 
despite significant appearance variations among distinct modalities, training on synthetic data from MRGen substantially improves segmentation accuracy across all organs. However, DualNorm, and segmentation models trained on data derived from CycleGAN and UNSB, yield unsatisfactory results.

\input{tables/ablation_study_vae_and_text_guided_generation}

\subsection{Ablation Studies}
\label{sec:ablation_studies}
To validate the effectiveness of our strategies and modules, we conduct comprehensive ablation experiments on both generation and downstream segmentation tasks, as follows.

\vspace{2pt}
\noindent \textbf{Generation.}
We assess autoencoder reconstruction and text-guided generation performance on MRGen-DB test set across various models, including:
(i) pretrained Stable Diffusion~(SDM), 
(ii) SDM finetuned on MRGen-DB~(SDM-ft), 
(iii) our model conditioned only on free-text modality labels~(MRGen-M), 
and 
(iv) our full MRGen with text prompts, 
incorporating modalities, attributes, regions, and organs information.
The reconstruction quality is assessed using PSNR, SSIM, and MSE between the reconstructed and original images. 
For generation, we use FID, along with image-to-image~(CLIP-I) and image-to-text~(CLIP-T) similarities between generated images and ground truth or modality labels, computed by BiomedCLIP~\cite{zhang2023biomedclip}.

As presented in Table~\ref{tab:ablation_study_vae_and_text_guided_generation}, we can draw the following observations:
(i) while SDM pretrained on natural images performs poorly on MRI, 
finetuning on MRGen-DB yields substantial improvement in both reconstruction and generation;
(ii) MRGen, with higher latent dimensions~(16 versus 4 of SDM) and the BiomedCLIP text encoder, achieves significantly better performance;
and 
(iii) our templated text prompts further enable MRGen to better distinguish distinct modalities, regions, and organs, leading to superior synthesis quality.
To this end, we employ high-capacity autoencoders trained on MRGen-DB, a text encoder pretrained on biomedical data, and clinically relevant templated text prompts to ensure accurate and realistic MRI synthesis.

\input{tables/ablation_study_segmentation}

\vspace{2pt}
\noindent \textbf{Segmentation.}
We train nnUNet with data generated by MRGen under various training and inference settings.
As shown in Table~\ref{tab:ablation_study_segmentation}, we have the following two observations:
(i) MRGen boosts segmentation performance with synthetic data, even without training on MR images from target domain, demonstrating its strong generalization capability to underrepresented and annotation-scarce modalities;
and
(ii) the inclusion of target-domain MR images~(without annotations) and the autofilter pipeline further improve performance by mitigating overfitting and selecting high-quality samples aligned with mask conditions.

%% file: tables/quantitative_results_generation.tex
\begin{table}[t]
\centering
\setlength{\tabcolsep}{1.5pt} 
\renewcommand{\arraystretch}{1.1} 
\resizebox{\linewidth}{!}{
\begin{tabular}{cc|cc|ccccc}
\toprule
\multirow{2}{*}{\centering \makecell{Source \\ Datset}} & 
\multirow{2}{*}{\centering \makecell{Source \\ Modality}} & 
\multirow{2}{*}{\centering \makecell{Target \\ Dataset}} & 
\multirow{2}{*}{\centering \makecell{Target \\ Modality}} & 
\multirow{2}{*}{\centering \makecell{Source \\ Domain}} & 
\multirow{2}{*}{\centering \makecell{DualNorm \\ \cite{zhou2022DualNorm}}} & 
\multirow{2}{*}{\centering \makecell{CycleGAN \\ \cite{CycleGAN2017}}} &
\multirow{2}{*}{\centering \makecell{UNSB \\ \cite{kim2024unpaired}}} &
\multirow{2}{*}{\centering \makecell{\textbf{MRGen} \\ \textbf{(Ours)}}} \\
& & & & & & & \\
\midrule
 CM. & T1 & CM. & T2-SPIR & 156.98 & 228.16 & 157.77 & 160.12 & \textbf{44.82} \\
 CM. & T2-SPIR & CM. & T1 & 156.98 & 261.97 & 188.91 & 141.15 & \textbf{60.79} \\
 MP. & T2 & MP. & ADC & 305.56 & 422.73 & 112.82 & 303.19 & \textbf{99.38} \\
 MP. & ADC & MP. & T2 & 305.56 & 416.31 & 190.82 & 346.60 & \textbf{123.55} \\
 PS. & T1 & PS. & T2 & 76.95 & 120.36 & 237.52 & 80.76 & \textbf{34.35} \\
 PS. & T2 & PS. & T1 & 76.95 & 126.27 & 89.26 & 76.91 & \textbf{58.90} \\
 LQ. & T1 & CM. & T2-SPIR & 109.46 & 281.56 & 182.62 & 193.05 & \textbf{88.76} \\
 CM. & T2-SPIR & LQ. & T1 & 109.46 & 246.73 & 260.06 & 192.80 & \textbf{106.45} \\
 MP. & ADC & PR. & T2 & 387.29 & 434.54 & 221.27 & 200.55 & \textbf{116.35} \\
 PR. & T2 & MP. & ADC & 387.29 & 365.10 & 140.72 & 252.64 & \textbf{88.43} \\
 \midrule
 \multicolumn{4}{c|}{\textbf{Average FID} $\downarrow$} & 207.25 & 290.37 & 178.18 & 194.78 & \textbf{82.18} \\
\bottomrule
\end{tabular}
}
\vspace{-3pt}
\caption{
\textbf{Quantitative Results~(FID) on Generation}.
Here, CM., MP., PS., LQ., and PR., denote CHAOS-MRI, MSD-Prostate, PanSeg, LiQA, and PROMISE12, respectively.
}
\label{tab:quantitative_results_generation}
\vspace{-12pt}
\end{table}


%% file: tables/quantitative_results_segmentation.tex
\begin{table*}[t]
\centering
\setlength{\tabcolsep}{3pt} 
\renewcommand{\arraystretch}{1.1} 
\resizebox{\linewidth}{!}{
\begin{tabular}{cc|cc|ccccccccc}
\toprule
\multirow{2}{*}{\centering \makecell{Source \\ Dataset}} & 
\multirow{2}{*}{\centering \makecell{Source \\ Modality}} & 
\multirow{2}{*}{\centering \makecell{Target \\ Dataset}} & 
\multirow{2}{*}{\centering \makecell{Target \\ Modality}} & 
\multirow{2}{*}{\centering \makecell{DualNorm \\ \cite{zhou2022DualNorm}}} & 
\multicolumn{4}{c}{UMamba~\cite{ma2024UMamba}} & 
\multicolumn{4}{c}{nnUNet~\cite{isensee2021nnUNet}} \\
\cline{6-9} \cline{10-13}
 & & & & & $\mathcal{D}_{s}$ & CycleGAN & UNSB & \textbf{MRGen} & $\mathcal{D}_{s}$ & CycleGAN & UNSB & \textbf{MRGen} \\
\midrule
\makecell{CHAOS-MRI} & 
T1 &
\makecell{CHAOS-MRI} & 
T2-SPIR &
14.00 & 4.02 & 10.58 & 20.56 & \textbf{67.35} & 6.90 & 7.58 & 14.03 & \underline{66.18}  \\
\makecell{CHAOS-MRI} &  
T2-SPIR &
\makecell{CHAOS-MRI} &   
T1 &
12.50 & 0.62 & 0.22 & 4.41 & \underline{57.24} & 0.80 & 1.38 & 6.44 & \textbf{58.10}  \\
\makecell{MSD-Prostate} &  
T2 &
\makecell{MSD-Prostate} &  
ADC &
1.43 & 0.19 & 45.06 & 45.77 & 52.58 & 5.52 & 40.92 & \underline{52.99} & \textbf{57.83}  \\
\makecell{MSD-Prostate} &  
ADC & 
\makecell{MSD-Prostate}  &  
T2 &
12.94 & 11.80 & \underline{62.00} & 36.47 & \textbf{64.05} & 22.20 & 57.06 & 38.39 & 61.95  \\
\makecell{PanSeg} &  
T1 & 
\makecell{PanSeg} &  
T2 &
0.21 & 0.38 & 2.13 & 3.08 & \underline{9.34} & 0.68 & 2.40 & 2.38 & \textbf{9.78}  \\
\makecell{PanSeg} & 
T2 & 
\makecell{PanSeg} &  
T1 &
0.11 & 0.27 & 5.08 & 4.46 & \underline{10.29} & 0.30 & 3.59 & 6.68 & \textbf{12.07} \\
\makecell{LiQA} &
T1 & 
\makecell{CHAOS-MRI} &  
T2-SPIR &
19.23 & 11.05 & 8.65 & 5.14 & \textbf{37.30} & 16.20 & 7.84 & 4.79 & \underline{31.73}  \\
\makecell{CHAOS-MRI} & 
T2-SPIR & 
\makecell{LiQA} & 
T1 &
31.09 & 10.33 & 41.22 & 28.52 & \underline{52.54} & 15.80 & 41.02 & 15.28 & \textbf{57.28} \\
\makecell{MSD-Prostate} & 
ADC &
\makecell{PROMISE12} & 
T2 &
1.43 & 23.71 & \underline{43.24} & \textbf{43.64} & 37.12 & 17.19 & 42.13 & 42.40 & 35.33  \\
\makecell{PROMISE12} & 
T2 &
\makecell{MSD-Prostate} & 
ADC &
9.84 & 21.75 & \underline{57.21} & 49.82 & 49.77 & 19.20 & \textbf{59.95} & 55.13 & 56.88 \\
\midrule
\multicolumn{4}{c|}{{\textbf{Average DSC score}}} & 10.28 & 8.41 & 27.54 & 24.19 & \underline{43.76} & 10.48 & 26.39 & 23.85 & \textbf{44.71}  \\
\bottomrule
\end{tabular}
}
\vspace{-3pt}
\caption{
\textbf{Quantitative Results~(DSC score) on Segmentation.}
Here, $\mathcal{D}_s$ denotes training with manually annotated source-domain data.
Results with the best and second best results are \textbf{bolded} and \underline{underlined}, respectively.
}
\label{tab:quantitative_results_segmentation}
\vspace{-9pt}
\end{table*}

%% file: tables/ablation_study_vae_and_text_guided_generation.tex
\begin{table}[t]
    \centering
    \small
    \setlength{\tabcolsep}{3pt} 
    \renewcommand{\arraystretch}{1.1}
    \centering
    \begin{tabular}{c|cccc}
        \toprule
        Model & SDM~\cite{SDM} & SDM-ft & MRGen-M & \textbf{MRGen} \\
        \midrule
        PSNR~$\uparrow$ & 31.32 & 35.65 & --- & \textbf{42.62} \\
        SSIM~$\uparrow$  & 0.989 & 0.996 & --- & \textbf{0.999}\\
        MSE~$\downarrow$  & 0.0037 & 0.0014 & --- & \textbf{0.0003} \\
        \midrule
        FID~$\downarrow$ & 249.24 & 91.48 & 41.82 & \textbf{39.63} \\
        CLIP-I~$\uparrow$ & 0.3151 & 0.6698 & 0.7512 & \textbf{0.8457} \\
        CLIP-T~$\uparrow$ & 0.1748 & 0.3199 & 0.3765 & \textbf{0.3777} \\
        \bottomrule
    \end{tabular}
    \vspace{-3pt}
    \caption{
    \textbf{Ablation on Reconstruction and Text-guided Generation.}
    Here, MRGen-M adopts the same autoencoder as MRGen.
    }
    \vspace{-9pt}
    \label{tab:ablation_study_vae_and_text_guided_generation}
\end{table}

%% file: tables/ablation_study_segmentation.tex
\begin{table}[t]
\centering
\setlength{\tabcolsep}{3pt} 
\renewcommand{\arraystretch}{1.1} 
    \resizebox{\linewidth}{!}{
    \begin{tabular}{c|cc|cccc}
    \toprule
    \multirow{2}{*}{\centering \makecell{Method}} & 
    \multirow{2}{*}{\centering \makecell{AutoFilter}} & 
    \multirow{2}{*}{\centering \makecell{$\mathcal{D}_t$ \\ Image}} & 
     \multicolumn{2}{c|}{CHAOS-MRI~\cite{kavur2021chaos}} & \multicolumn{2}{c}{MSD-Prostate~\cite{antonelli2022MSD}} \\ 
    \cline{4-5} \cline{6-7}
     & & &
     T1 $\rightarrow$ T2S. & T2S. $\rightarrow$ T1 & T2 $\rightarrow$ ADC & ADC $\rightarrow$ T2 \\
    \midrule
    nnUNet~\cite{isensee2021nnUNet} & \XSolidBrush & \XSolidBrush & 6.90 & 0.80 & 5.52 & 22.20 \\
    \midrule
    \multirow{4}{*}{\centering \makecell{nnUNet \\ (\textbf{MRGen})}} 
    & \XSolidBrush & \XSolidBrush & 16.53 & 15.10 & 39.90 & 18.92 \\
    & \Checkmark & \XSolidBrush & 22.30 & 20.27 & 42.79 & 25.34 \\
    & \XSolidBrush & \Checkmark & 30.16 & 29.01 & 49.04 & 40.89 \\
    & \Checkmark & \Checkmark & \textbf{66.18} & \textbf{58.10} & \textbf{57.83} & \textbf{61.95} \\
    \bottomrule
    \end{tabular}
    }
    \vspace{-6pt}
    \caption{
    \textbf{Ablation on Segmentation Performance~(DSC score).}
    }
\label{tab:ablation_study_segmentation}
\vspace{-12pt}
\end{table}

%% file: sec/5_conclusion.tex
\section{Conclusion}
This paper explores generative models for controllable MRI generation, particularly to facilitate training segmentation models for underrepresented modalities lacking mask annotations. 
To support this, we curate a large-scale radiology image-text dataset, \textbf{MRGen-DB}, featuring detailed modality labels, attributes, regions, and organ information, with a subset of organ mask annotations.
Built on this, our diffusion-based data engine, \textbf{MRGen}, synthesizes MR images of various annotation-scarce modalities conditioned on text prompts and masks.
Comprehensive evaluations across diverse modalities demonstrate that MRGen effectively improves segmentation performance on unannotated modalities by producing high-quality synthetic data. 
We believe this will offer new insights into addressing the scarcity of medical data and annotations, holding clinical significance.

%% file: sec/6_acknowledgement.tex
\vspace{-0.3cm}
\section*{Acknowledgments}
\vspace{-0.1cm}
This work is supported by the National Key R \& D Program of China (No. 2022ZD0161400).

%% file: sec/X_suppl.tex
\onecolumn
{
    \centering
    \Large
    \textbf{MRGen: Segmentation Data Engine for Underrepresented MRI Modalities}\\
    \vspace{0.5em} Appendix \\
    \vspace{1.0em}
}

\appendix
{
  \hypersetup{linkcolor=black}
  \tableofcontents
}

\clearpage

\section{Preliminaries on Diffusion Models}
\label{sec:preliminaries}
\noindent \textbf{Diffusion Models}~\cite{ho2020ddpm} are a class of deep generative models that convert Gaussian noise into structured data samples through an iterative denoising process. 
These models typically comprise a forward diffusion process and a reverse denoising process.

Specifically, the forward diffusion process progressively introduces Gaussian noise into an image~($\mathbf{x}_0$) via a Markov process over $T$ steps.
Let $\mathbf{x}_t$ represent the noisy image at step $t$. 
The transition from $\mathbf{x}_{t-1}$ to $\mathbf{x}_t$ can be formulated as:
\begin{align*}
    q(\mathbf{x}_t | \mathbf{x}_{t-1}) = \mathcal{N}(\mathbf{x}_t; \sqrt{1 - \beta_t}\mathbf{x}_{t-1}, \beta_t \mathbf{I})
\end{align*}
Here, $\beta_t \in (0, 1)$ represents pre-determined hyperparameters that control the variance at each step.
By defining $\alpha_t = 1 - \beta_t$ and $\bar{\alpha}_t = \prod_{i=1}^{t} \alpha_i$, the properties of Gaussian distributions and the reparameterization trick allow for a refined expression:
\begin{align*}
    q(\mathbf{x}_t | \mathbf{x}_0) = \mathcal{N}(\mathbf{x}_t; \sqrt{\bar{\alpha}_t}\mathbf{x}_0, (1 - \bar{\alpha}_t)\mathbf{I})    
\end{align*}
This insight provides a concise expression for the forward process with Gaussian noise $\boldsymbol{\epsilon}$ as: $\mathbf{x}_t = \sqrt{\bar{\alpha}_t}\mathbf{x}_0 + \sqrt{1 - \bar{\alpha}_t} \boldsymbol{\epsilon}$.

Diffusion models also encompass a reverse denoising process that reconstructs images from noise.
A UNet-based model~\cite{ronneberger2015UNet} is typically utilized to learn the reverse diffusion process $p_\theta$, represented as:
\begin{align*}
    p_{\theta}(\mathbf{x}_{t-1} | \mathbf{x}_t) = \mathcal{N}(\mathbf{x}_t; \mu_{\theta}(\mathbf{x}_t, t), \Sigma_{\theta}(\mathbf{x}_t, t))   
\end{align*}
Here, $\mu_{\theta}$ represents the predicted mean of the Gaussian distribution, derived from the estimated noise $\boldsymbol{\epsilon}_{\theta}$ as:
\begin{align*}
    \mu_{\theta}(\mathbf{x}_t, t) = \frac{1}{\sqrt{\alpha_t}}(\mathbf{x}_t - \frac{1 - \alpha_t}{\sqrt{1 - \bar{\alpha}_t}}\boldsymbol{\epsilon}_{\theta}(\mathbf{x}_t, t))
\end{align*}

Building on this foundation, \textbf{Latent Diffusion Models}~\cite{SDM} adopt a Variational Autoencoder~(VAE~\cite{kingma2013VAE}) to project images into a learned, compressed, low-dimensional latent space.
The forward diffusion and reverse denoising processes are then performed on the latent codes~($\mathbf{z}$) within this latent space, significantly reducing computational cost and improving efficiency.

\section{Details of MRGen-DB \& Synthetic Data}
\label{sec:more_details_of_data}
This section provides additional details about our curated \textbf{MRGen-DB} dataset.
In Sec.~\ref{sec:automatic_annotations}, we elaborate on the implementation details of the automatic annotation pipeline;
and in Sec.~\ref{sec:dataset_statistics}, we present more comprehensive data statistics.
Moreover, in Sec.~\ref{sec:synthetic_data_statistics}, we provide statistics on the MRGen-synthesized data used for downstream segmentation models training.

\subsection{Automatic Annotations}
\label{sec:automatic_annotations}
We employ an automated annotation pipeline to annotate our MRGen-DB dataset, ensuring that the templated text prompts contain sufficient and clinically relevant information to distinguish distinct modalities, regions, and organs.
This process primarily consists of two precise and controllable components: human body region classification and modality explanation, which will be detailed as follows.

\vspace{2pt}
\noindent \textbf{Region classification.}
Considering the wide range and variability of abdominal imaging, we adopt the off-the-shelf BiomedCLIP~\cite{zhang2023biomedclip} image encoder to encode all 2D slices, and the BiomedCLIP text encoder to encode predefined text descriptions of six abdominal regions. 
Based on the cosine similarity between the image and text embeddings, the 2D slices are classified into one of the six categories, including {\em Upper Thoracic Region}, {\em Middle Thoracic Region}, {\em Lower Thoracic Region}, {\em Upper Abdominal Region}, {\em Lower Abdominal Region}, and {\em Pelvic Region}. 
For text encoding, we use a templated text prompt as input:

\begin{mdframed}[backgroundcolor=gray!10, linewidth=0pt]
{\em 
    This is a radiology image that shows \$region\$ of a human body, and probably contains \$organ\$.
}
\end{mdframed}

Here, \$region\$ and \$organ\$ represent the items in the following list:

\begin{mdframed}[backgroundcolor=gray!10, linewidth=0pt]
{\em 
    (region, organ)  = [
        (`Upper Thoracic Region', `lung, ribs and clavicles'),
        (`Middle Thoracic Region', `lung, ribs and heart'),
        (`Lower Thoracic Region', `lung, ribs and liver'),
        (`Upper Abdominal Region', `liver, spleen, pancreas, kidney and stomach'),
        (`Lower Abdominal Region', `kidney, small intestine, colon, cecum and appendix'),
        (`Pelvic Region', `rectum, bladder, prostate/uterus and pelvic bones')
    ]
}
\end{mdframed}

\vspace{2pt}
\noindent \textbf{Modality explanation.}
To capture the correlations and distinctions among various modality labels, we leverage GPT-4~\cite{achiam2023gpt4} to generate free-text descriptions detailing the signal intensities of {\em fat}, {\em muscle}, and {\em water} for each modality label.
This helps the model better understand the imaging characteristics of distinct modalities. 
The prompt we use is as follows:

\begin{mdframed}[backgroundcolor=gray!10, linewidth=0pt]
{\em 
    As a senior doctor and medical imaging researcher, please help me map radiological imaging modalities to the signal intensities of fat, muscle, and water, as well as their corresponding brightness levels. Please provide the answer in the following format:  
    fat \{\} signal, muscle \{\} signal, water \{\} signal, fat \{\}, muscle \{\}, water \{\}.
    Now, tell me the attributes of \$modality\$.
}
\end{mdframed}

To ensure reliability and accuracy, we have randomly and uniformly sampled approximately 2\%~(5K out of 250K) of region annotations and 20\%~(60 out of $\sim$300) of modality attribute annotations for manual verification, achieving high accuracies of 95.33\% and 91.67\%.
Furthermore, the effectiveness in downstream tasks also validates the quality of automatic annotations.

\subsection{Dataset Statistics}
\label{sec:dataset_statistics}
In this section, we present more detailed statistics about our curated MRGen-DB dataset, including the unannotated image-text pairs from {\em Radiopaedia}\footnote{radiopaeida.org}, as well as the mask-annotated data sourced from various open-source datasets.

\begin{figure}[h]
  \centering
  \includegraphics[width=\textwidth]{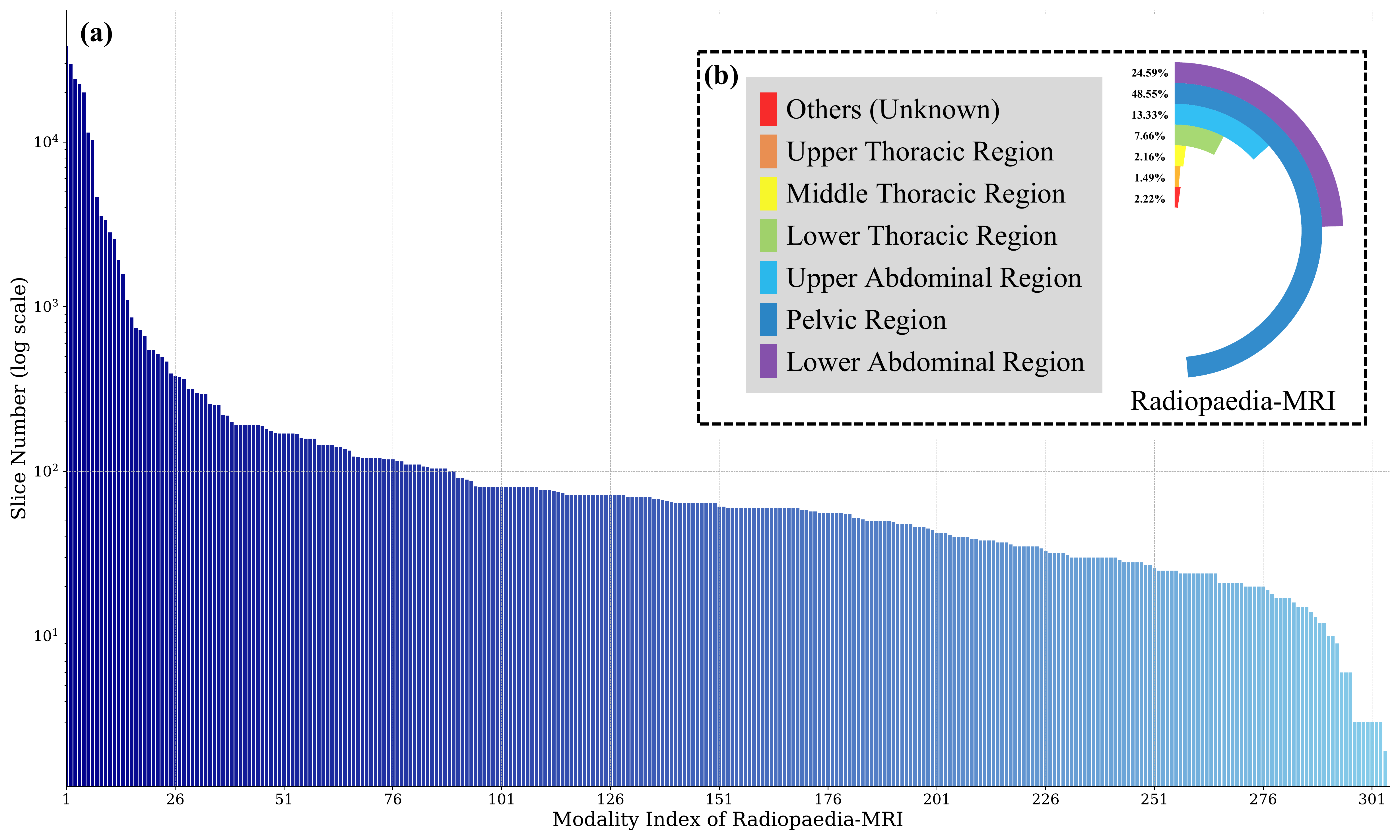} \\
  \vspace{-6pt}
  \caption{
  \textbf{Data Statistics of {\em Radiopaedia-MRI}.}
  (a) Distribution of slice counts across various modalities in {\em Radiopaedia-MRI};
  (b) Proportional distribution of slices across different regions in {\em Radiopaedia-MRI}.
  }
    \vspace{-6pt}
 \label{fig:radiopaedia_statistics}
\end{figure}

\noindent \textbf{Data without mask annotations.}
For the image-text pairs from {\em Radiopaedia-MRI}, which are used for training the autoencoder and text-guided generation, we allocate 1\% of the data as a test set to evaluate reconstruction and generation performance, maximizing the amount of data available for pretraining. 
As a result, 202,988 samples are used for training, and the test set consists of 2,051 samples.
We conduct a statistical analysis of the distribution of modalities in {\em Radiopaedia-MRI}, as presented in Figure~\ref{fig:radiopaedia_statistics}~(a).
The free-text modality labels cover approximately 300 categories, providing a diverse set of MRI modalities that form a crucial foundation for MRGen to learn text-guided generation and expand its mask-conditioned generation capabilities towards modalities originally lacking mask annotations.
Furthermore, the distribution of images across different regions in {\em Radiopaedia-MRI} is presented in Figure~\ref{fig:radiopaedia_statistics}~(b).

\vspace{2pt}
\noindent \textbf{Data with mask annotations.}
Following the SAT~\cite{zhao2023one}, we split the data with mask annotations into training and test sets, as detailed in Table~\ref{tab:detailed_dataset_statistics}.
For dataset pairs comprising different datasets, we use their shared organs as the segmentation targets.

\input{tables/detailed_dataset_statistics}

\subsection{Synthetic Data Statistics}
\label{sec:synthetic_data_statistics}
This section presents the statistics of target-domain training samples synthesized by MRGen across various experimental settings, as presented in Table~\ref{tab:synthetic_data_statistics}.
Concretely, we use mask annotations from the entire source-domain dataset~(including both training and test sets) as input conditions to generate target-domain images, forming image-mask training pairs.
Exceptions include:
(i) for the MSD-Prostate~\cite{antonelli2022MSD} dataset, where images of T2 and ADC modalities have already been registered, we restrict inputs to the source-domain training set to prevent data leakage; 
and
(ii) for dataset pairs with CHAOS-MRI-T1~\cite{kavur2021chaos} as the target domain, each source-domain mask is used twice to synthesize both T1-InPhase and T1-OutofPhase data.
This setup is consistent across all baselines.
Additionally, with our proposed autofilter pipeline, MRGen generates 20 candidate images per mask and selects the top two that meet the predefined thresholds. 
If no samples satisfy the thresholds, all thresholds will be relaxed by 0.10, and the sample of the highest quality is chosen, ensuring full exploitation of source-domain masks.
Otherwise, all low-quality generated samples are discarded to avoid noisy data.

\input{tables/synthetic_data_statistics}

\section{Implementation Details}
In this section, we will provide a comprehensive explanation of the implementation details discussed in the paper.
Concretely, Sec.~\ref{sec:preprocessing_augmentation} describes the preprocessing and augmentation strategies applied to the training data. 
Sec.~\ref{sec:autofilter_pipeline} elaborates on the details of the autofilter pipeline. 
Finally, Sec.~\ref{sec:baselines} outlines the implementation details of various baselines.

\subsection{Preprocessing \& Augmentation}
\label{sec:preprocessing_augmentation}
\noindent \textbf{Data preprocessing.}
To ensure consistency across data from various sources and modalities, we apply tailored preprocessing strategies as follows:
(i) For data from {\em Radiopaedia-MRI}, the images are directly rescaled to the range [0, 1];
(ii) For MR images with mask annotations, intensities are clipped to the 0.5 and 99.5 percentiles and rescaled to [0, 1].
After normalization, all data are subsequently rescaled to [-1, 1] for training various components of MRGen, including the autoencoder, diffusion UNet, and mask condition controller.
For training downstream segmentation models, images are rescaled to [0, 255] and saved in `.png' format, followed by the official preprocessing configurations of nnUNet~\cite{isensee2021nnUNet} and UMamba~\cite{ma2024UMamba}.

\vspace{2pt}
\noindent \textbf{Data augmentation.}
During autoencoder training, we apply random data augmentations to images with a 20\% probability.
These augmentations include horizontal flipping, vertical flipping, and rotations of 90$^\circ$, 180$^\circ$, 270$^\circ$.
In contrast, no data augmentations are applied during the training of the diffusion UNet and mask condition controller.
Our preliminary experiments show that MRGen remains robust to uneven data distribution; we therefore do not explicitly adopt data balancing in training.
For segmentation models, we adhere to the default data augmentation strategies provided by nnUNet~\cite{isensee2021nnUNet} and UMamba~\cite{ma2024UMamba}.

\subsection{Autofilter Pipeline}
\label{sec:autofilter_pipeline}
When deploying our proposed data engine, MRGen, to synthesize training data for segmentation models, we adopt the off-the-shelf SAM2-Large~\cite{ravi2024sam2} model to perform automatic interactive segmentation on generated images, with the mask conditions as spatial prompts.
Empirically, we observe that SAM2 consistently segments images based on their contours, guided by the provided spatial prompts.
Concretely, it produces high-quality pseudo mask annotations for images with contours closely matching mask conditions, while performing poorly for synthesized images that deviate significantly from mask conditions.
This characteristic allows our pipeline to automatically filter out samples faithful to the condition masks and discard erroneous ones, thus ensuring the quality of synthesized image-mask pairs.
Here, we elaborate on more implementation details of this automatic filtering pipeline, particularly focusing on the generation of MR images that encompass multiple organs.

Specifically, we begin by defining the following thresholds: confidence threshold~($\tau_{\mathrm{conf}}$), IoU score threshold~($\tau_{\mathrm{IoU}}$), average confidence threshold~($\bar{\tau}_{\mathrm{conf}}$), and average IoU threshold~($\bar{\tau}_{\mathrm{IoU}}$).
Both the controlling mask~($\mathcal{M'}_t$) and the generated image~($\mathcal{I'}_t$) are fed into SAM2.
For each organ mask $\mathcal{M'}^i_t$ in $\mathcal{M'}_t$, SAM2 will output a segmentation map with a confidence score~($s^i_{\mathrm{conf}}$), which is then used to calculate the IoU score~($s^i_{\mathrm{IoU}}$) against $\mathcal{M'}^i_{t}$.
For each generated sample~($\mathcal{I'}_t$), it is regarded to be high-quality and aligned with the mask condition if the following conditions are satisfied: $\{s^{i}_{\mathrm{IoU}} \geq \tau_{\mathrm{IoU}}, s^{i}_{\mathrm{conf}} \geq \tau_{\mathrm{conf}} \, | \, \forall i \}$, 
and
$\{\bar{s}_{\mathrm{IoU}} \geq \bar{\tau}_{\mathrm{IoU}}$, $\bar{s}_{\mathrm{conf}} \geq \bar{\tau}_{\mathrm{conf}} \}$.
Otherwise, the sample will be discarded.

For each conditional mask, we synthesize 20 image candidates and select the best two that satisfy the predefined thresholds.
Across all experiments, the thresholds are set as follows:
$\tau_{\mathrm{IoU}} = 0.70$, $\tau_{\mathrm{conf}} = 0.80$, $\bar{\tau}_{\mathrm{IoU}} = 0.80$, and $\bar{\tau}_{\mathrm{conf}} = 0.90$.

\subsection{Baselines}
\label{sec:baselines}
In this section, we introduce the implementation details of representative baselines and discuss other relevant methods by category.
Concretely, we first consider the most related ones, including augmentation-based and translation-based methods.

\vspace{2pt}
\noindent \textbf{Augmentation-based methods.}
These approaches~\cite{zhou2022DualNorm, hu2023devil, su2023rethinking, ouyang2022causality, xu2022adversarial, chen2023treasure} typically rely on mixing multi-domain training data or employing meticulously designed data augmentation strategies.
Here, we consider the representative one, DualNorm~\cite{zhou2022DualNorm}.
Following its official implementation, we apply random non-linear augmentation on each source-domain image, to generate a source-dissimilar training sample, and train the dual-normalization model.
All preprocessing steps, network architectures, and training strategies adhere to the official recommendations, with the exception that images are resized to 512 $\times$ 512, consistent with other methods.
Notably, we evaluate DualNorm on all slices in the test set, offering a more rigorous evaluation compared to the official code, which only considers slices with segmentation annotations.

\vspace{2pt}
\noindent \textbf{Translation-based methods.}
These methods~\cite{sasaki2021unit, kim2024unpaired, phan2023structure, kim2024adaptive} are commonly inspired by CycleGAN~\cite{CycleGAN2017}; therefore, we compare with open-source CycleGAN~\cite{CycleGAN2017}, UNSB~\cite{kim2024unpaired}, and MaskGAN~\cite{phan2023structure}.
We follow their official implementations and training strategies across all experimental settings. 
Subsequently, source-domain images are translated into the target domain and paired with the source-domain masks to create paired samples for training downstream segmentation models.

Moreover, we have also explored other approaches leveraging the progress of generative models.

\vspace{2pt}
\noindent \textbf{Generation-based methods.}
Existing medical generation models~\cite{zhan2024medm2g, guo2024maisi, hamamci2024generatect, dorjsembe2024BrainMRISyn, wang2024self} still struggle with complex and challenging abdominal MRI generation.
For instance, MAISI~\cite{guo2024maisi} and Med-DDPM~\cite{dorjsembe2024BrainMRISyn} are tailored for CT and brain MRI synthesis, respectively. 
To adapt to our task, we finetune MAISI~\cite{guo2024maisi} on our data, as a generation-based baseline.

Additionally, we consider other methods aimed at addressing our focused challenge, {\em i.e.}, segmenting MR images of underrepresented modalities lacking mask annotations.
These include few-shot learning approaches, general-purpose segmentation models, and methods incorporating oracle inputs as performance references.
Notably, these approaches, to varying degrees, rely on manually annotated target-domain segmentation masks or external datasets.
Thus, they should be regarded as references only, rather than fair comparisons with the aforementioned methods and our MRGen.

\vspace{2pt}
\noindent \textbf{Few-shot methods.}
Specifically, we compare with a few-shot nnUNet~\cite{isensee2021nnUNet}~(pre-trained on source-domain data and finetuned on 5\% target-domain manually annotated data), as well as UniVerSeg~\cite{butoi2023universeg} with its official implementation and checkpoint.

\vspace{2pt}
\noindent \textbf{General segmentation models.}
We adopt the official code and checkpoint of TotalSegmentor-MRI~\cite{d2024TotalSegmentorMRI}, which has been trained on extensive manually annotated data and diverse modalities, as a strong general-purpose segmentation baseline.

\vspace{2pt}
\noindent \textbf{Models with oracle inputs.}
We include SAM2-Large~\cite{ravi2024sam2} as a reference for interactive semi-automatic segmentation, using randomly perturbed oracle boxes as prompts. 
To simulate the error introduced by manual intervention, the oracle boxes are randomly shifted at each corner, by up to 8\% of the image resolution, following MedSAM~\cite{MedSAM}.
Segmentation results are derived in a slice-by-slice and organ-by-organ manner: For each slice with mask annotations, we simulate box prompts for each annotated organ individually.
Finally, we also include nnUNet~\cite{isensee2021nnUNet} trained exclusively on the target-domain mask-annotated dataset~($\mathcal{D}_t$) as an oracle reference, reflecting the performance upper bound with sufficient annotated data.

\section{More Experiments}
In this section, we present additional experimental results to demonstrate the superiority of our proposed data engine. 
First, in Sec.~\ref{sec:in-domain_generation}, we showcase quantitative and qualitative results of in-domain generation. 
Next, in Sec.~\ref{sec:more_quantitative_results}, we present quantitative comparisons with more baselines, further confirming the effectiveness and necessity of our proposed data engine.
Then, in Sec.~\ref{sec:extension_to_more_modalities}, we present extra promising application prospects~(cross-protocol generation) on paired CT and MRI datasets.
Finally, in Sec.~\ref{sec:more_quanlitative_results}, we provide extra qualitative results to validate the accuracy and flexibility of the generated outputs. 

\subsection{In-domain Generation}
\label{sec:in-domain_generation}
Our proposed data engine not only synthesizes images for target modalities lacking mask annotations but also maintains controllable generation capabilities within the source domains. 
Moreover, as presented in Table~\ref{tab:in-domain_generation}, downstream segmentation models trained exclusively on synthetic source-domain data can achieve performance comparable to those trained on real, manually-annotated data.
This offers a feasible solution to address concerns about medical data privacy.

\begin{figure}[t]
  \centering
  \includegraphics[width=.9\textwidth]{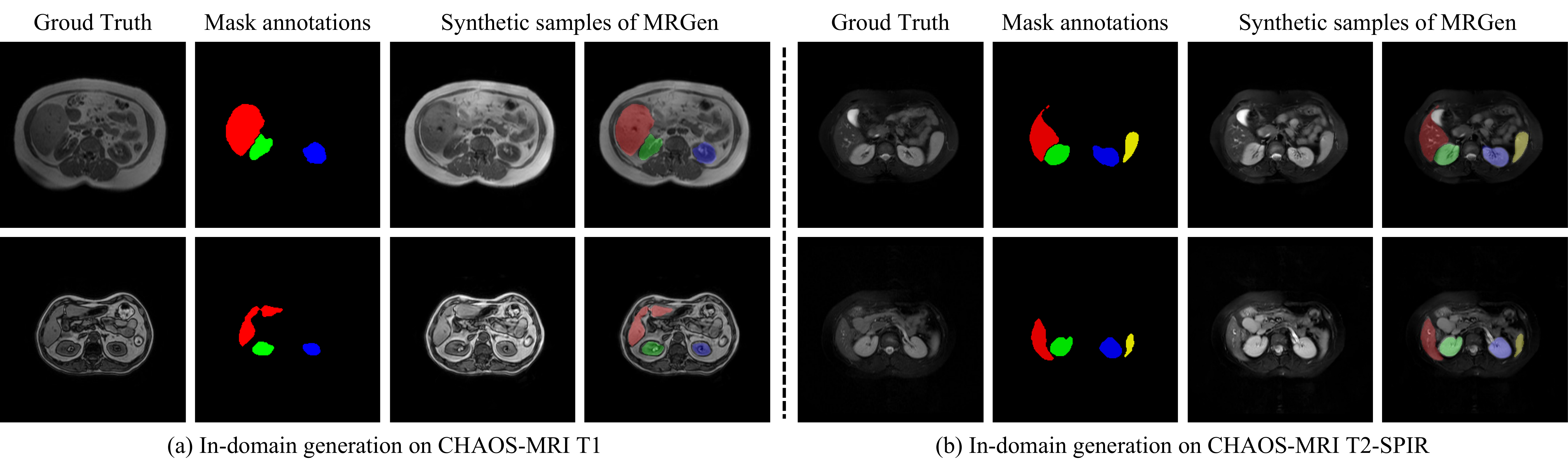} \\
  \vspace{-6pt}
  \caption{
    \textbf{Qualitative Results of In-domain Generation.}
  }
    \vspace{-15pt}
 \label{fig:in-domain_generation}
\end{figure}

\input{tables/in-domain_generation}

Moreover, we provide visualizations of in-domain generation in Figure~\ref{fig:in-domain_generation}, qualitatively demonstrating that our MRGen can reliably perform controllable generation of a large number of samples within the training domain with mask annotations.

\input{tables/more_quantitative_results}

\subsection{More Quantitative Results}
\label{sec:more_quantitative_results}
In this section, we compare MRGen with additional baseline methods on two typical cross-modal dataset pairs from MRGen-DB by evaluating the performance of downstream segmentation models, as detailed in the main text. 
Concretely, for both translation-based and generation-based methods, we assess the performance of nnUNet~\cite{isensee2021nnUNet} trained on data generated by these methods.
As depicted in Table~\ref{tab:more_quantitative_results}, we further analyze the relevant baselines by category, as follows.

\vspace{2pt}
\noindent \textbf{Augmentation-based methods.}
Limited to relying on carefully crafted augmentation strategies, DualNorm~\cite{zhou2022DualNorm} fails to model nonlinear visual discrepancies among distinct modalities, leading to poor cross-modality segmentation performance.

\vspace{2pt}
\noindent
\textbf{Translation-based methods.}
While CycleGAN~\cite{CycleGAN2017}, UNSB~\cite{kim2024unpaired}, and MaskGAN~\cite{phan2023structure} excel at contour preservation, they often suffer from model collapse when learning complex modality transformations, resulting in suboptimal performance.

\vspace{2pt}
\noindent \textbf{Generation-based models.} 
Despite finetuned on our dataset, the performance of MAISI~\cite{guo2024maisi} is still poor, which we attribute to its lack of \textbf{modality-conditioning} capability. 
This limitation hinders its ability to support \textbf{cross-modality generation}, and consequently, makes it struggle to synthesize target-domain samples for training segmentation models.

\vspace{2pt}
\noindent \textbf{Few-shot methods.}
While few-shot nnUNet~\cite{isensee2021nnUNet} and UniverSeg~\cite{butoi2023universeg} benefit from partial target-domain annotations, MRGen-boosted models outperform without requiring any such annotations, showcasing practical feasibility in clinical scenarios.

\vspace{2pt}
\noindent
\textbf{General segmentation models.}
TotalSegmentor-MRI~\cite{d2024TotalSegmentorMRI} works well on certain datasets/modalities~(\textbf{likely already included during training}), but it still performs poorly or even fails on others.
This significantly limits its practicality in complex clinical scenarios, especially when dealing with underrepresented modalities with diverse imaging characteristics.

\vspace{2pt}
\noindent \textbf{Models with oracle inputs.}
Although SAM2~\cite{ravi2024sam2} with perturbed oracle boxes as prompts exhibits impressive zero-shot segmentation capabilities, our MRGen-boosted models still outperform it, trailing only the oracle nnUNet trained directly on target-domain annotated data.
Moreover, as a semi-automatic method, SAM2's reliance on high-quality spatial prompts and manual intervention limits its scalability and applicability, while MRGen offers a fully automated, end-to-end solution.

Overall, MRGen provides a robust, fully automated approach for challenging cross-modality segmentation by producing high-quality synthetic data, with no need for any target-domain mask annotations and proving highly suitable for clinical applications.
For computational efficiency, we primarily focus on comparing MRGen with some representative baselines, DualNorm~\cite{zhou2022DualNorm}, CycleGAN~\cite{CycleGAN2017} and UNSB~\cite{kim2024unpaired}, across more dataset pairs in the main text for a comprehensive evaluation.

\input{tables/ct_to_mri}

\subsection{Extension to More Modalities}
\label{sec:extension_to_more_modalities}
Considering that evaluation on truly rare modalities is difficult due to limited ground truth annotations, we simulate such scenarios by restricting models from accessing target-modality labels in our experiments.
Here, we also explore cross-modality synthesis~(from CT to MRI) with AMOS22~\cite{ji2022amos}, MSD-Liver~\cite{antonelli2022MSD}, and CHAOS-MRI~\cite{kavur2021chaos} datasets to further demonstrate MRGen's potential for broader cross-protocol generation, as depicted in Table~\ref{tab:ct_to_mri}.

\subsection{More Qualitative Results}
\label{sec:more_quanlitative_results}
In this section, we provide qualitative visualizations of more datasets, covering both image generation and segmentation.

\begin{figure}[p]
  \centering
  \includegraphics[width=.95\textwidth]{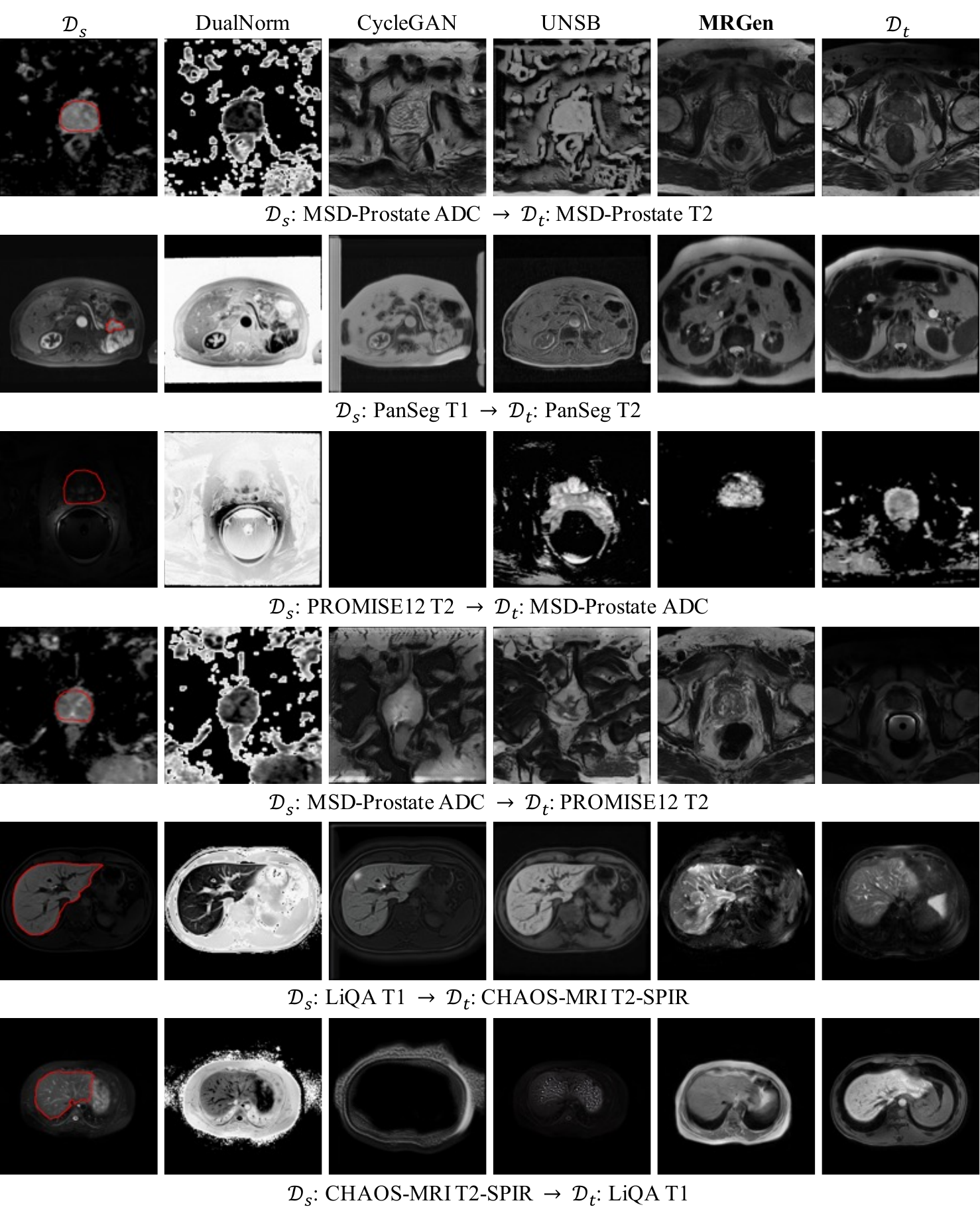} \\
  \vspace{-8pt}
  \caption{
  \textbf{More Qualitative Results of Controllable Generation.}
  We present images from source domains~($\mathcal{D}_s$) and target domains~($\mathcal{D}_t$) for reference.
  Here, specific organs are contoured with colors: \textcolor{red}{prostate} in MSD-Prostate and PROMISE12 datasets, and \textcolor{red}{pancreas} in PanSeg dataset, and \textcolor{red}{liver} in LiQA and CHAOS-MRI datasets.
  }
    \vspace{-4pt}
 \label{fig:more_qualitative_results_generation}
\end{figure}

\vspace{2pt}
\noindent \textbf{Image generation.}
We present extra visualizations of controllable generation on target modalities lacking mask annotations in Figure~\ref{fig:more_qualitative_results_generation}, which demonstrate that our MRGen can effectively generate high-quality samples based on masks across various datasets and modalities, facilitating the training of downstream segmentation models towards these challenging scenarios.

\begin{figure}[p]
  \centering
  \includegraphics[width=\textwidth]{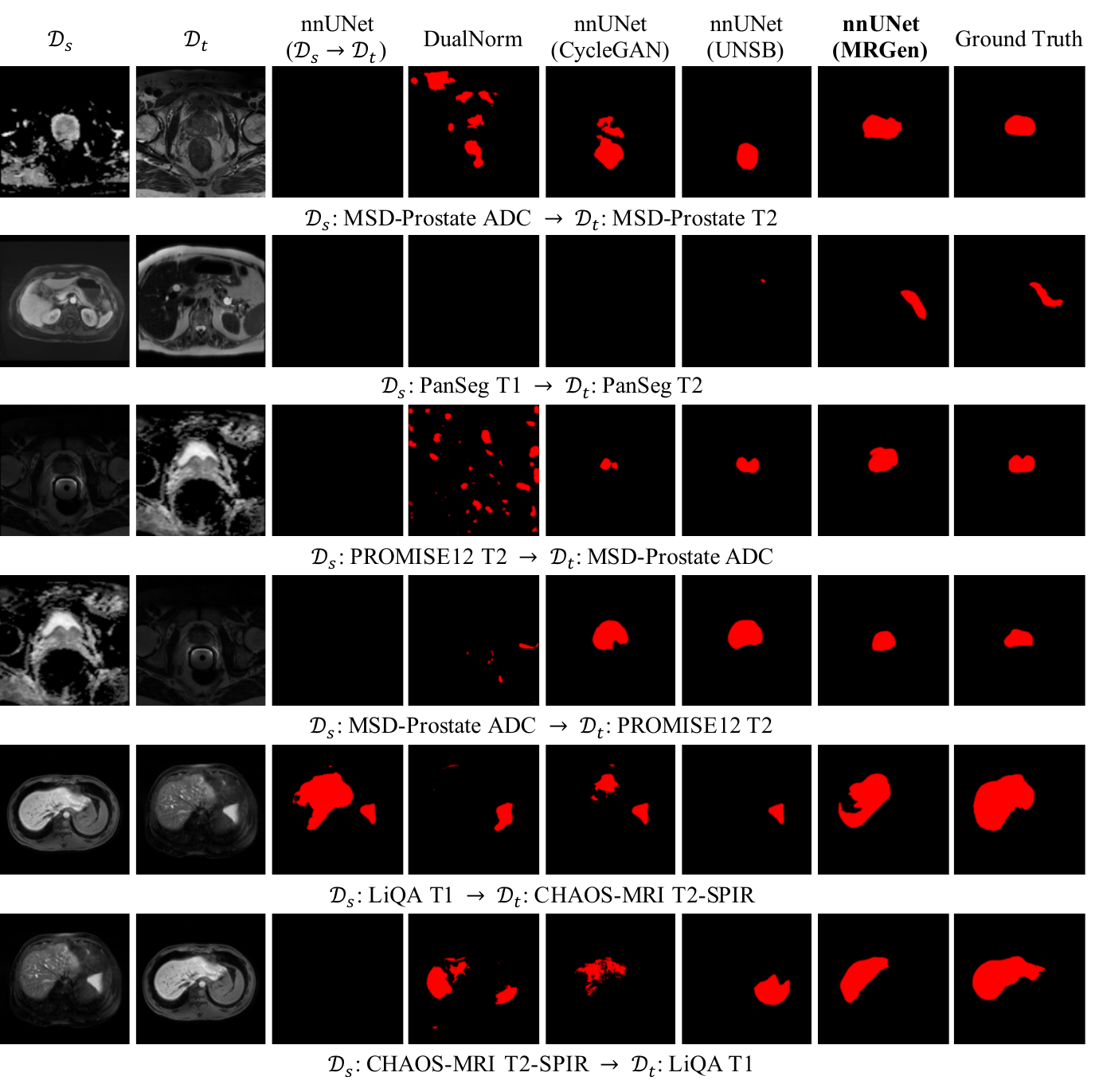} \\
  \vspace{-8pt}
  \caption{
  \textbf{More Qualitative Results on Segmentation towards Unannotated Modalities.} 
  Significant imaging differences between source-domain~($\mathcal{D}_s$) and target-domain~($\mathcal{D}_t$) make segmentation on target domains~($\mathcal{D}_t$) extremely challenging.
  Here, specific organs are highlighted with colors: \textcolor{red}{prostate} in MSD-Prostate and PROMISE12, \textcolor{red}{pancreas} in PanSeg, and \textcolor{red}{liver} in LiQA and CHAOS-MRI datasets.
  }
    \vspace{-4pt}
 \label{fig:more_qualitative_results_segmentation}
\end{figure}

\vspace{2pt}
\noindent \textbf{Image segmentation.}
As presented in Figure~\ref{fig:more_qualitative_results_segmentation}, we provide more visualizations of segmentation models trained using synthetic data on modalities that originally lack mask annotations.
This validates that the samples generated by MRGen can effectively assist in training segmentation models, achieving impressive performance in previously unannotated scenarios.

\begin{figure}[h]
  \centering
  \includegraphics[width=.9\textwidth]{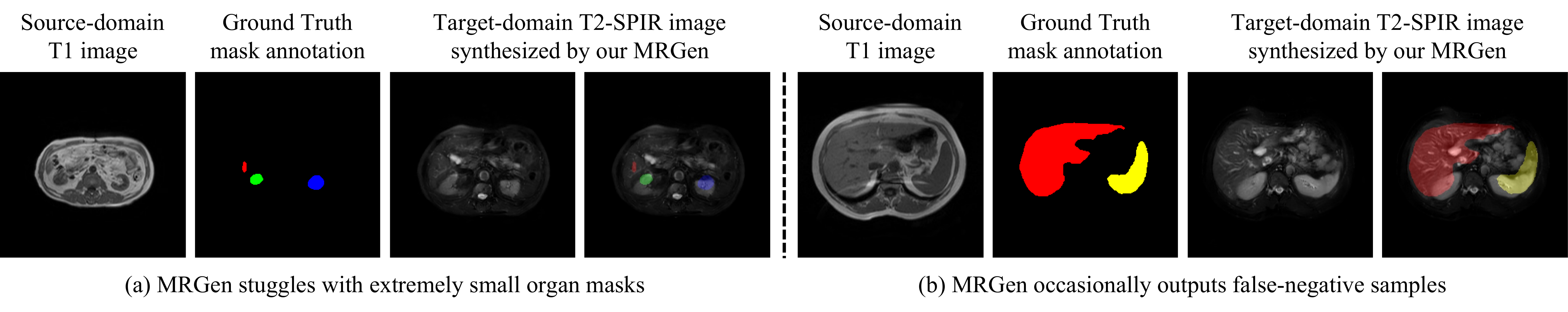} \\
  \vspace{-6pt}
  \caption{
  \textbf{Failure Cases Analysis.} 
  Our proposed MRGen is not without limitations: 
  (a) it may struggle to handle extremely small organ masks; 
  (b) it occasionally produces false-negative samples, such as the unexpected synthesis of kidneys in the given example.
  }
    \vspace{-12pt}
 \label{fig:bad_case}
\end{figure}

\section{Limitations \& Future Works}
\subsection{Limitations}
Our proposed data engine, MRGen, is not without its limitations.
Specifically, MRGen encounters difficulties when generating conditioned on extremely small organ masks and occasionally produces false-negative samples.

\vspace{2pt}
\noindent \textbf{Extremely small organ masks.}
The morphology of the same organ, such as the {\em liver} or {\em spleen}, can vary significantly across different slices of a 3D volume, resulting in significant variability in their corresponding masks. 
Furthermore, the distribution of these masks is often imbalanced, with extremely small masks being relatively rare.
When generating in the latent space, these masks are further downsampled, leading to unstable generation quality, as depicted in Figure~\ref{fig:bad_case} (a).
A feasible solution to mitigate this issue is to increase the amount of data with mask annotations, thereby improving the model’s robustness.

\vspace{2pt}
\noindent \textbf{False-negative samples.}
Another challenge arises from the varying number of organs on each slice. 
For instance, one slice may contain {\em liver}, {\em kidneys}, and {\em spleen}, while another may include only {\em liver} and {\em spleen}. 
This variability causes MRGen to occasionally generate targets not specified in the mask condition. 
As depicted in Figure~\ref{fig:bad_case} (b), {\em kidneys} are unexpectedly synthesized, despite not being included in the mask conditions, leading to false negatives during the training of downstream segmentation networks.
A feasible solution is to design a more robust data filtering pipeline to filter false-negative samples, and simple manual selection can also serve as a quick and effective method to remove the non-compliant samples.

\begin{figure}[htpb]
  \centering
  \includegraphics[width=.88\textwidth]{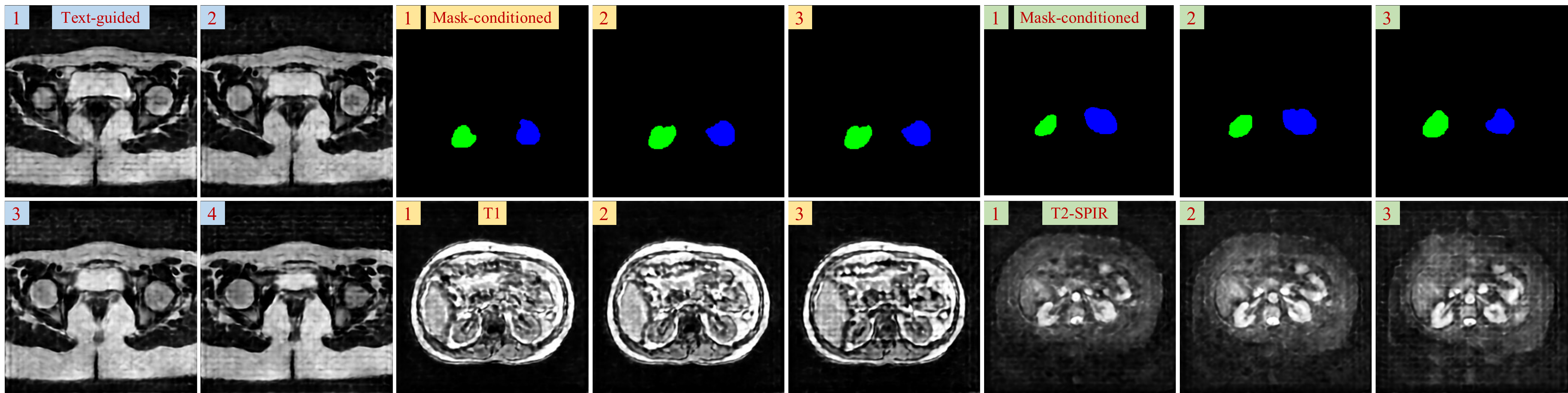} \\
  \vspace{-3pt}
  \caption{
    \textbf{The validation of 3D extension feasibility of MRGen on $256 \times 256 \times 16$ volumes.} 
  }
  \label{fig:3D}
  \vspace{-12pt}
\end{figure}

\subsection{Future Works}
Due to limited computational resources, we validate our data engine on 2D slices, with trained segmentation models able to process 3D volumes slice-by-slice.
However, our idea can be seamlessly extended to 3D volume generation with more computing in the future to further advance cross-modality segmentation performance.
Here, we provide a preliminary validation on $256 \times 256 \times 16$ volumes, as depicted in Figure~\ref{fig:3D}.
While the results are not fully optimized due to limited computations, they already demonstrate promising \textbf{inter-slice consistency}, indicating the feasibility of extending MRGen to 3D synthesis.

Moreover, to address the aforementioned limitations of MRGen, we propose several directions for future improvement:
(i) Constructing more comprehensive and richly annotated datasets, such as incorporating more annotated MRI data, to enhance the model’s ability to effectively utilize mask conditions;
(ii) Designing finer-grained and efficient generative model architectures to improve generation efficiency and accuracy, particularly for small-volume organs;
and 
(iii) Developing a more robust data filtering pipeline to reliably select high-quality samples that meet the requirements of downstream tasks.

\clearpage

%% file: tables/detailed_dataset_statistics.tex
\begin{table}[t]
    \centering
    \small
    \renewcommand{\arraystretch}{1.1} 
    \resizebox{\textwidth}{!}{
    \begin{tabular}{lllcccccc}
    \toprule
    \multirow{2}{*}{\centering Dataset} & \multirow{2}{*}{\centering \centering Organs} & \multirow{2}{*}{\centering \centering Modality} & \multicolumn{3}{c}{ Train} & \multicolumn{3}{c}{Test} \\
    \cline{4-6} \cline{7-9}
    & & & \#~Vol. & \#~Slc. & \#~Slc. w/~mask & \#~Vol. & \#~Slc. & \#~Slc. w/~mask \\
    \midrule
    \multirow{2}{*}{\centering PanSeg~\cite{zhang2024PanSeg}} & \multirow{2}{*}{\centering Pancreas} & T1-weighted & 309 & 14,656 & 5,961 & 75 & 3,428 & 1,400 \\
    & & T2-weighted & 305 & 12,294 & 5,106 & 77 & 2,982 & 1,312 \\
    \midrule
    \multirow{2}{*}{\centering MSD-Prostate~\cite{antonelli2022MSD}} & \multirow{2}{*}{\centering Prostate} & T2-weighted & 26 & 492 & 100 & 6 & 110 & 83 \\
    & & ADC & 26 & 492 & 100 & 6 & 110 & 83 \\
    \midrule
    \multirow{2}{*}{CHAOS-MRI~\cite{kavur2021chaos}} & Liver, Right Kidney, & T1-weighted & 32 & 1,018 & 770 & 8 & 276 & 230 \\
    & Left Kidney, Spleen & T2-SPIR & 16 & 503 & 388 & 4 & 120 & 104 \\
    \midrule
    PROMISE12~\cite{litjens2014PROMISE12} & Prostate & T2-weighed & 40 & 1,137 & 645 & 10 & 240 & 133 \\
    \midrule
    LiQA~\cite{liu2024LiQA} & Liver & T1-weighted & 24 & 1,718 & 1,148 & 6 & 467 & 298 \\
    \midrule
    \textbf{Total} & / & / & \textbf{778} & \textbf{36,710} & \textbf{14,218} & \textbf{192} & \textbf{7,733} & \textbf{3,643} \\
    \bottomrule
    \end{tabular}
    }
    \vspace{-3pt}
    \caption{
    \textbf{Details of Segmentation-annotated Datasets in MRGen-DB.}
    Here, \#~Vol. represents the number of 3D Volumes, \#~Slc. denotes the number of 2D slices, and \#~Slc.~w/~mask indicates the number of 2D slices with mask annotations.
    }
    \label{tab:detailed_dataset_statistics}
    \vspace{-12pt}
\end{table}



%% file: tables/synthetic_data_statistics.tex
\begin{table*}[ht]
    \centering
    \setlength{\tabcolsep}{3pt} 
    \renewcommand{\arraystretch}{1.1}
    \resizebox{.75\linewidth}{!}{
    \begin{tabular}{cc|cc|cc}
    \toprule
    Source Dataset & Source Modality & Target Dataset & Target Modality & \#~Slices~($\mathcal{D}_s$) & \#~Synthetic Data \\
    \midrule
     CHAOS-MRI & T1 & CHAOS-MRI & T2-SPIR & 1,294 & 433 \\
     CHAOS-MRI & T2-SPIR & CHAOS-MRI & T1 & 607 & 1,118 \\
     MSD-Prostate & T2 & MSD-Prostate & ADC & 492 & 775 \\
     MSD-Prostate & ADC & MSD-Prostate & T2 & 492 & 745 \\
     PanSeg & T1 & PanSeg & T2 & 18,084 & 2,160 \\
     PanSeg & T2 & PanSeg & T1 & 15,276 & 2,215 \\
     LiQA & T1 & CHAOS-MRI & T2-SPIR & 2,185 & 2,267 \\
     CHAOS-MRI & T2-SPIR & LiQA & T1 & 607 & 636 \\
     MSD-Prostate & ADC & PROMISE12 & T2 & 602 & 742 \\
     PROMISE12 & T2 & MSD-Prostate & ADC & 1,377 & 1,077 \\
    \bottomrule
    \end{tabular}
    }
    \vspace{-3pt}
    \caption{
    \textbf{Synthetic Data Statistics.}
        Here, \#~Slices~($\mathcal{D}_s$) denotes the number of source-domain samples under each experimental setting, which serves as input for translation-based baselines.
        Moreover, \#~Synthetic Data represents the total volume of data generated by MRGen.
    }
\label{tab:synthetic_data_statistics}
\vspace{-15pt}
\end{table*}


%% file: tables/in-domain_generation.tex
\begin{table}[h]
\centering
\small
\setlength{\tabcolsep}{3pt} 
\renewcommand{\arraystretch}{1.1} 
    \begin{tabular}{c|cc|cc|ccc}
    \toprule
    \multirow{2}{*}{\centering \makecell{Dataset}}
    & \multirow{2}{*}{\centering \makecell{Source \\ Modality}} 
    & \multirow{2}{*}{\centering \makecell{Target \\ Modality}}
    & \multicolumn{2}{c|}{$\mathcal{D}_s$}
    & \multicolumn{3}{c}{$\mathcal{D}_t$} 
    \\
    \cline{4-5} \cline{6-8}
    & & & $\mathcal{D}_s$ & \textbf{MRGen} & $\mathcal{D}_s$ & \textbf{MRGen} & $\mathcal{D}_t$
     \\
    \midrule
     \multirow{2}{*}{\centering CHAOS-MRI~\cite{kavur2021chaos}} 
     & T1 & T2-SPIR & \cellcolor{gray!20}90.60 & \textbf{88.14} & 4.02 & \textbf{67.35} & \cellcolor{gray!20}83.90 \\
     & T2-SPIR & T1 & \cellcolor{gray!20}83.90 & \textbf{82.06} & 0.62 & \textbf{57.24} & \cellcolor{gray!20}90.60 \\
    \bottomrule
    \end{tabular}
    \vspace{-3pt}
    \caption{
        \textbf{In-domain \& Cross-domain Augmentation Results~(DSC score) on Segmentation}.
        We compare the performance of nnUNet~\cite{isensee2021nnUNet} trained on real data versus synthetic data generated by MRGen in both the source domain~($\mathcal{D}_s$) and target domain~($\mathcal{D}_t$).
    }
    \label{tab:in-domain_generation}
    \vspace{-6pt}
\end{table}

%% file: tables/more_quantitative_results.tex
\begin{table}[ht]
    \centering
    \setlength{\tabcolsep}{3pt} 
    \renewcommand{\arraystretch}{1.1} 
    \resizebox{\linewidth}{!}{
    \begin{tabular}{c|c|c|cccccccc|cc|ccc}
    \toprule
    \multirow{2}{*}{\centering \makecell{Dataset}} &
    \multirow{2}{*}{\centering \makecell{Source \\ Modality}} & 
    \multirow{2}{*}{\centering \makecell{Target \\ Modality}} & 
    \multirow{2}{*}{\centering \makecell{DualNorm}} & 
    \multicolumn{7}{c|}{nnUNet} & & & & &
     \\
    \cline{5-11}
     & & & & $\mathcal{D}_s$ & \textbf{MRGen} & CycleGAN & UNSB & MaskGAN & MAISI & Few-shot & \multirow{-2}{*}{\centering \makecell{UniVerSeg}} & \multirow{-2}{*}{\centering \makecell{TS-MRI}} & 
     \multirow{-2}{*}{\centering \makecell{Oracle \\ Box}} & 
    \multirow{-2}{*}{\centering SAM2} & 
    \multirow{-2}{*}{\centering \makecell{nnUNet\\ $\mathcal{D}_t$}}
     \\
    \midrule
    \multirow{2}{*}{\makecell{CHAOS-MRI}} & \makecell{T1} & \makecell{T2-SPIR} & 14.00 & 6.90 & \textbf{66.18} & 7.58 & 14.03 & \underline{32.73} & 3.34 & \cellcolor{gray!20} 52.00 & \cellcolor{gray!20} 48.91 & \cellcolor{gray!20} 80.64 & \cellcolor{gray!20} 45.45 & \cellcolor{gray!20} 53.12 & \cellcolor{gray!20} 83.90 \\
     & \makecell{T2-SPIR} & \makecell{T1} & \underline{12.50} & 0.80 & \textbf{58.10} & 1.38 & 6.44 & 1.89 & 3.11 & \cellcolor{gray!20} 53.82 & \cellcolor{gray!20} 52.79 & \cellcolor{gray!20} 77.09 & \cellcolor{gray!20} 43.48 & \cellcolor{gray!20} 51.94 & \cellcolor{gray!20} 90.60 \\
     \midrule
    \multirow{2}{*}{\makecell{MSD-Prostate}} & \makecell{T2} & \makecell{ADC} & 1.43 & 5.52 & \textbf{57.83} & 40.92 & \underline{52.99} & 29.14 & 9.15 & \cellcolor{gray!20} 20.28 & \cellcolor{gray!20} 0.0 & \cellcolor{gray!20} 0.0 & \cellcolor{gray!20} 61.50 & \cellcolor{gray!20} 65.39 & \cellcolor{gray!20} 82.35 \\
    & \makecell{ADC} & \makecell{T2} & 12.94 & 22.20 & \textbf{61.95} & \underline{57.06} & 38.39 & 5.98 & 6.94 & \cellcolor{gray!20} 29.38 & \cellcolor{gray!20} 53.90 & \cellcolor{gray!20} 0.0 & \cellcolor{gray!20} 61.07 & \cellcolor{gray!20} 66.40 & \cellcolor{gray!20} 89.80 \\
    \midrule
    \multicolumn{3}{c|}{{\textbf{Average DSC score}}} & 10.22 & 8.86 & \textbf{61.02} & 26.74 & \underline{27.96} & 17.44 & 5.64 & \cellcolor{gray!20} 38.87 & \cellcolor{gray!20} 38.90 & \cellcolor{gray!20} 39.43 & \cellcolor{gray!20} 52.88 & \cellcolor{gray!20} 59.21 & \cellcolor{gray!20} 86.66 \\
    \bottomrule
    \end{tabular}
    }
    \vspace{-3pt}
    \caption{
    \textbf{More Quantitative Results~(DSC score) on Segmentation.}
    The best and second-best performances are \textbf{bolded} and \underline{underlined}, respectively.
    Notably, the results marked with a \setlength{\fboxsep}{1pt}\colorbox{gray!20}{gray background} indicate that the corresponding methods may have accessed target-modality annotated data during extensive training~({\em e.g.}, UniVerSeg, TotalSegmentor-MRI~(TS-MRI)), utilized oracle inputs as prompts~({\em e.g.}, Oracle Box, SAM2), or even been directly trained on target-modality annotated data~({\em e.g.}, nnUNet~(Few-shot), nnUNet~($\mathcal{D}_t$)).
    Consequently, these approaches do not represent a fully fair comparison with others, and are primarily included as performance references.
    }
    \label{tab:more_quantitative_results}
    \vspace{-12pt}
\end{table}

%% file: tables/ct_to_mri.tex
\begin{table}[htp]
    \centering
    \setlength{\tabcolsep}{3pt} 
    \renewcommand{\arraystretch}{1.1} 
    \resizebox{.7\linewidth}{!}{
    \begin{tabular}{c|c|ccccccc}
    \toprule
    \multirow{2}{*}{\centering \makecell{Source \\ Domain}} & 
    \multirow{2}{*}{\centering \makecell{Target \\ Domain}} & 
    \multirow{2}{*}{\centering \makecell{DualNorm}} &
    \multicolumn{3}{c}{nnUNet} & 
    \multicolumn{3}{c}{UMamba}
     \\
    \cline{4-9}
     & & & $\mathcal{D}_s$ & CycleGAN & \textbf{MRGen} & $\mathcal{D}_s$ & CycleGAN & \textbf{MRGen}  \\ 
    \midrule
    \makecell{AMOS22~(CT)} & \makecell{CHAOS~(T2)} & 19.78 & 0.11 & 6.75 & \underline{22.50} & 0.05 & 8.06 & \textbf{26.73} \\
    \makecell{AMOS22~(CT)} & \makecell{CHAOS~(T1)} & 16.09 & 8.88 & 52.49 & \underline{56.23} & 3.19 & 43.21 & \textbf{60.53}  \\
    \makecell{MSD-Liver~(CT)} & \makecell{CHAOS~(T2)} & 1.58 & 3.12 & 10.14 & \underline{38.67} & 1.65 & 11.06 & \textbf{40.93} \\
    \bottomrule
    \end{tabular}
    }
    \vspace{-3pt}
    \caption{\textbf{Quantitative Results~(DSC score) on Cross-protocol settings~(from CT to MRI).}}
    \label{tab:ct_to_mri}
    \vspace{-12pt}
\end{table}